\documentclass[11pt]{article}

\usepackage[final]{acl}

\usepackage{times}
\usepackage{latexsym}
\usepackage{placeins}
\usepackage{needspace}
\usepackage{multirow}
\usepackage{amsfonts}
\usepackage{hyperref}
\usepackage[a-1b]{pdfx}
\usepackage{amsmath}
\usepackage{subfigure}
\usepackage{afterpage}
\usepackage{enumitem}
\usepackage{wrapfig}
\usepackage{lipsum,graphicx}
\usepackage{svg}
\usepackage{bbding}
\usepackage{pdfpages}
\usepackage{pgfplots}  
\usepackage{amsfonts}

\usepackage[T1]{fontenc}
\usepackage[greek,english]{babel}

\usepackage{microtype}
\usepackage[normalem]{ulem}

\usepackage{booktabs}

\usepackage[scaled=0.8]{beramono}

\usepackage[normalem]{ulem}

\title{Dialect Normalization using Large Language Models and Morphological Rules}

 \author{Antonios Dimakis\textsuperscript{$\alpha$,$\beta$}, John Pavlopoulos\textsuperscript{$\alpha$,$\gamma$}, Antonios Anastasopoulos\textsuperscript{$\alpha$,$\delta$} \\
         \textsuperscript{$\alpha$}Archimedes, Athena Research Center, Greece \\ \textsuperscript{$\beta$}Department of Informatics and Telecommunications, NKUA \\ \textsuperscript{$\gamma$}Department of Informatics, Athens University of Economics and Business, Greece \\ \textsuperscript{$\delta$}Department of Computer Science, George Mason University \\ \texttt{andimakis@di.uoa.gr, ipavlopoulos@aueb.gr, antonis@gmu.edu}}

\begin{document}
\maketitle
\begin{abstract}

Natural language understanding systems struggle with low-resource languages, including many dialects of high-resource ones. Dialect-to-standard normalization attempts to tackle this issue by transforming dialectal text so that it can be used by standard-language tools downstream. In this study, we tackle this task by introducing a new normalization method that combines rule-based linguistically informed transformations and large language models (LLMs) with targeted few-shot prompting, without requiring any parallel data. We implement our method for Greek dialects and apply it on a dataset of regional proverbs, evaluating the outputs using human annotators. We then use this dataset to conduct downstream experiments, finding that previous results regarding these proverbs relied solely on superficial linguistic information, including orthographic artifacts, while new observations can still be made through the remaining semantics.\footnote{We publicly release all code and datasets produced for this work: \url{https://github.com/andhmak/rule_dialnorm}}

\end{abstract}

\section{Introduction}

Natural language processing has long struggled with lower-resource language varieties, including geographic varieties of higher-resource standardized ones \cite{10.1145/3712060}. According to members of such language communities, who are usually also speakers of the equivalent standard, natural language understanding (NLU) of dialectal text is much more important than language generation (NLG) into the local variety \cite{blaschke-etal-2024-dialect}.

This variation in demand highlights the significance of transforming dialectal text into a standard variety while maintaining as much of the original meaning as possible, which is known as dialect-to-standard normalization. That is because by improving our methods in this area we will be able to apply modern NLU techniques to a vast array of formerly neglected varieties through models trained on data of their related standard languages.

In this work, we introduce a novel method for normalizing dialectal data into a standard variety. Our proposed method first applies morphological rules, specified based on dialect-specific linguistic prior knowledge, and then feeds the preprocessed input to an LLM along with dialect-specific shots. This second step enhances the input with sentences exhibiting those facultative dialectal features which are not addressable only with the first step.

\begin{figure}[t]
    \centering
    \small
    \begin{tabular}{@{}c@{}c@{}}
        \multicolumn{2}{l}{\textbf{Source Dialectal Sentence}} \\
        \multicolumn{2}{l}{src: \foreignlanguage{greek}{Ου Θεός κι ου γείτονας.}} \\
        \multicolumn{2}{l}{\ \ \ \ \ \ \  /u theós ki u jítonas/} \\
        \multicolumn{2}{l}{\ \ \ \ \ \ \ \textit{God and the neighbour.}} \\
        \cmidrule{1-1}
        \multicolumn{2}{l}{\textbf{Baseline Normalization}} \\
        \multicolumn{1}{l}{$\rightarrow$: \foreignlanguage{greek}{Ούτε ο Θεός, ούτε ο γείτονας.}} & \textcolor{red}{$\mathbf{\mathcal{X}}$}\\
        \multicolumn{2}{l}{\ \ \ \ \ \ \  /úte o theós, úte o jítonas/} \\
        \multicolumn{2}{l}{\ \ \ \ \ \ \  \textit{Neither God nor the neighbour.}} \\
        \multicolumn{2}{l}{\textbf{Proposed Rule-Enhanced Method}} \\
        \multicolumn{1}{l}{$\rightarrow$ \foreignlanguage{greek}{Ο Θεός και ο γείτονας.}} & \textcolor{OliveGreen}{$\mathbf{\checkmark}$}\\
        \multicolumn{2}{l}{\ \ \ \ \ \ \  /o theós ke o jítonas/} \\
        \multicolumn{2}{l}{\ \ \ \ \ \ \  \textit{God and the neighbour.}}
    \end{tabular}
    \caption{Predictable phonological changes (/o/$\rightarrow$/u/)
    in Northern Greek dialects make the definite article ``o'' appear closer to Standard ``\foreignlanguage{greek}{ούτε}'' (/úte/, \textit{neither}). We combine LLMs with rule-based normalization to better understand dialectal sentences.}
    \label{fig:example}
\end{figure}

We implement the language-specific parts of this procedure for a set of Greek dialects represented in a large dataset of regional proverbs. 
An example of our method compared to simple prompting for one of the proverbs in our dataset is shown in Figure~\ref{fig:example}.
We then experiment with two different LLMs and ablate the rule-based step, using human annotators.
We thus produce a new normalized dataset in Standard Modern Greek, which we use in downstream tasks: 
first, we replicate prior research using the newly-standardized proverbs to ascertain whether the previous results depended on the semantics or on the now-removed linguistic peculiarities of each variety and its transcription method.

Additionally, we conduct further experiments showcasing the usability of our dataset for obtaining semantic, non-dialectally-colored insights into a set of originally dialectal texts.

\noindent In short, we make the following contributions:

\begin{itemize}[noitemsep,nolistsep,leftmargin=*]
  \item We propose a new method for normalizing dialectal speech, using a pipeline of rule-based transformations followed by an LLM with a few dialect-specific examples.
  \item As a proof-of-concept, we implement the linguistic rules for Greek dialects and run our pipeline on a dataset of Greek proverbs, producing a normalized dataset of regional proverbs, validated using human annotations.
  \item We show that previous observations into the original dataset could have been influenced by dialectal linguistic features, which disappear in the standardized text, while new, mainly semantic-based insights are possible.
\end{itemize}

\section{Related Work}

Previous work has been carried out in the area of dialect normalization, targeting specific varieties \cite{arabic, finnish, swiss}, as well as more generalized approaches \cite{general}.

Recently, pretrained multilingual LLMs have proven useful in such tasks, especially when fine-tuned on parallel dialectal-standard data \cite{alam-anastasopoulos-2025-normalization}.
These kinds of parallel datasets are in some way or another required in all these past techniques in order to train specialized models. In contrast, our technique eliminates this requirement by leveraging LLMs' tendency to treat unseen dialectal features as noise, combined with the exploitation of linguistic knowledge of the dialects in question and as few as three parallel sentences for few-shot prompting. This makes our approach viable even for use cases such as the one we explore where there are practically no parallel text data available.

\citet{pavlopoulos-etal-2024-towards} introduced a machine-actionable dataset of Greek proverbs, comprising over 100,000 proverb variants, each originating from one of 134 unique locations across Greece. An exploration of the spatial distribution of proverbs showed that the most widespread proverbs come from the mainland while the least come primarily from the islands. Using the latter, then, they showed that text geolocation/geocoding \cite{hovy-purschke-2018-capturing, han-etal-2016-twitter, chakravarthi-etal-2021-findings-vardial, ramponi-casula-2023-geolingit} can be accurate for specific locations, and that conventional machine learning algorithms operating on stylistic features outperformed transfer learning.
We argue, however, that relying on the superficial linguistic features of the original (non-normalized) text, instead of semantic ones, makes it hard to determine shared semantics or any (possibly deeper) cultural connections across different regions.

\section{Methodology}
\label{sec:methodology}

Our normalization method consists of two steps. First, we preprocess our inputs using a rule-based procedure. Then, we pass the previous step's output to an LLM with few-shot prompting.

\paragraph{Part 1: Rule-based normalization (RBN)}
RBN is achieved by string replacements of specific character sequences according to the linguistic features of each dialect compared to the standard.
We divide the Greek dialects into three groups, following established literature \cite{dialects}: Northern, Southern and Pontic, according to their features, and use different transformation rules for each group. The dialects' specific distribution among these groups is described in Appendix~\ref{sec:groups}, and indicative examples of string replacements are in Appendix~\ref{sec:rules}.
The amount of linguistic knowledge required is roughly what would be present in a dialect's comparative grammar, in our case amounting to 14 string replacement rules.

\paragraph{Part 2: Few-shot prompting}
Our prompts are designed to guide the model to perform our desired task while also providing the LLM with the necessary linguistic information, which is otherwise difficult to encode using rules.
First, we include the name of the region our text is sourced from (especially helpful if the model has seen relevant data during pre-training).
Second, we provide instructions to only change the dialect, so that it conforms with the standard, without affecting the style of the text. Otherwise, we notice that LLMs tend to view dialectal features as signs of informality, and therefore produce overly formal text when not explicitly directed not to. Similarly, they seem to replace vocabulary existing in both the dialect and the standard with alternatives. Hence, we also instruct for lexical terms to only be replaced when they are absent from the standard.
Finally, we provide a few examples of the task being performed successfully, specifically selected to display dialectal features not encoded in the previous step. The full prompt used per dialectal group is provided in Appendix~\ref{sec:templates}.

\section{Normalization Experiments}
\paragraph{Dataset}
We perform our experiments on the dataset provided by \citet{pavlopoulos-etal-2024-towards}, specifically on the balanced corpus, containing 500 proverbs from each geographic location, which was also used for their experiments.

\paragraph{Models}

For the LLM-based part of our normalization method, we use 
\textbf{GPT-4o} \cite[\texttt{gpt-4o-2024-11-20};][]{openai2024gpt4ocard} as well as the \textbf{Llama 3.1-70B} \cite{grattafiori2024llama3herdmodels}.
Overall we explore four different setups:

\begin{enumerate}[noitemsep,nolistsep,leftmargin=*]
    \item \textbf{\texttt{GPT 3s+RBN}} uses GPT-4o and follows the entire pipeline as described in Section~\ref{sec:methodology};
    \item \textbf{\texttt{GPT 3s}} only uses the 3-shot prompting method, using a different prompt according to the group of the input dialect, skipping RBN;
    \item \textbf{\texttt{Llama 3s+RBN}} uses Llama 3 and also follows the entire pipeline; and
    \item \textbf{\texttt{Llama 9s}} uses Llama 3 and skips both RBN and the division into dialectal groups, providing all three parallel examples of all three dialectal groups in every prompt.
\end{enumerate}

\paragraph{Human evaluation}
We employed three native Greek speakers to evaluate a subset of the normalized proverb dataset.
For each sentence, normalized with each of the four setups, they were instructed to provide a score from 1 to 5 on two axes. One was ``form'', referring to how well the normalized sentence was stripped of its dialectal features and rendered into fluent Standard Modern Greek. The other was ``meaning'', referring to how well the original meaning of the dialectal sentence, including its style, was preserved in the normalized one.
For each of these two axes, they were also asked to choose the best normalized sentence out of the four, with ties only allowed in case of identical output strings.
We derived various statistical measures guaranteeing the reliability of the annotations, inter-annotator agreement, and statistical significance. Table~\ref{tab:icc_summary} depicts the high Intraclass Correlation Coefficients (ICC) calculated on the annotations for each model on both axes, and more detailed results can be found in Appendix~\ref{sec:stats}.

\begin{table}[h]
    \centering
    \small
    \setlength{\tabcolsep}{10pt}
    \begin{tabular}{lcc}
    \toprule
    \textbf{Model} & \textbf{Form ICC} & \textbf{Meaning ICC} \\
    \midrule
    \verb|GPT 3s+RBM|   & 0.884  & 0.783  \\
    \verb|GPT 3s| & 0.934  & 0.893  \\
    \verb|Llama 3s+RBN| & 0.790  & 0.910  \\
    \verb|Llama 9s| & 0.888  & 0.875  \\
    \bottomrule
    \end{tabular}
    \caption{ICC for ``form'' and ``meaning'' ratings per model. Values closer to 1 indicate better correlation.}
    \label{tab:icc_summary}
\end{table}

\paragraph{Results}
Table~\ref{tab:norm_quality}, on the left, depicts the average human evaluation from 1 (worst) to 5 (best), based on form and meaning. The percentage each setup was assessed by human evaluators as the best (for form and meaning) is shown on the right. 
\verb|GPT 3s+RBN| was the best in form and meaning, followed by \verb|GPT 3s|, \verb|Llama 3s+RBN| and \verb|Llama 9s|. Differences among models are more prominent when explicitly asking for a preferred output.

\begin{table}[t]
    \centering
    \small
    \begin{tabular}{@{}l|cccc@{}}
    \toprule
        & \multicolumn{2}{c}{\textbf{Normalization}} & \multicolumn{2}{c}{\textbf{Percentage}}  \\
        & \multicolumn{2}{c}{\textbf{Quality (out of 5)}} & \multicolumn{2}{c}{\textbf{Best (\%)}}  \\
        \textbf{Model} & \textbf{Form} & \textbf{Meaning} & \textbf{Form} & \textbf{Meaning} \\
        \midrule
        GPT 3s+RBN & \textbf{4.68} & \textbf{4.62} & \textbf{88.3} & \textbf{91.5} \\
        GPT 3s & 4.46 & 4.26 & 66.8 & 68.6 \\
        Llama 3s+RBN & 3.1 & 3 & 16.7 & 13.5 \\
        Llama 9s & 2.52 & 2.34 & 9.3 & 9.7 \\
    \bottomrule
    \end{tabular}
    \caption{Average human evaluation, from 1 to 5 (best), regarding the form and meaning of the output per setup. GPT 3s+RBN is the best (left) and its output is the preferred normalization about 90\% of the time (right).}
    \label{tab:norm_quality}
\end{table}

\section{Downstream tasks}

\subsection{Text Geocoding/Geolocation}\label{ssec:geolocation}
We replicated the experiments of \citet{pavlopoulos-etal-2024-towards}, by using their corpus but normalizing it with our best performing approach (\verb|GPT 3s+RBN|). This includes training models for: (a) predicting the region label for each proverb without providing any further geographic information; and (b) for predicting the geographic coordinates using regression, after providing each region's exact location.
After removing any non-semantic and dialectal information (i.e., normalizing), we find that geolocation methods fail. This finding verifies the hypothesis of \citet{pavlopoulos-etal-2024-towards} about predictions being based on linguistic information. 

\noindent\textbf{In the classification geolocation task}, using the non-normalized data, the best model reaches an average F$_1$ score of 0.33, with that of specific regions being as high as 0.81. Using normalized data, however, the best model reaches only 0.13 (see Table~\ref{tab:class_summary}), with no region going above 0.35.

\noindent\textbf{In the regression geolocation task}, performance decreased less, going from an average root mean square error of 2.9 to 3.2 (see Table~\ref{tab:regr_summary}). The full results can be found in Appendix~\ref{sec:results}.

\paragraph{Semantic or superficial?} Compared to the results of the non-normalized analysis, models trained on our normalized data rely more on semantic, rather than on superficial linguistic features, such as transcription conventions. For instance, the top four terms guiding the best geolocation model (trained on non-normalized data) Southwards comprise different transcriptions of the conjunction \foreignlanguage{greek}{και} (kai, and). That is, they are phonologically affected by the Southern phenomenon of velar palatalization. However, when the same model is trained on the normalized version, it utilizes mainly semantically meaningful content words.

\begin{table}[h]
    \centering
    \small
    \setlength{\tabcolsep}{6pt}
    \begin{tabular}{lcc}
    \toprule
    \textbf{Model} & \textbf{Dialectal F$_1$} & \textbf{Normalized F$_1$} \\
    \midrule
    Logistic Regression  & \textbf{0.29} & 0.12 \\
    SVM                  & \textbf{0.33} & 0.13 \\
    K Nearest Neighbors  & \textbf{0.23} & 0.11 \\
    Random Forest        & \textbf{0.25} & 0.13 \\
    \bottomrule
    \end{tabular}
    \caption{Average F$_1$ per model for region (label) prediction, using the original dialectal dataset and our normalized one. Classification is harder after removing superficial linguistic information.}
    \label{tab:class_summary}
\end{table}

\begin{table}[h]
    \centering
    \small
    \begin{tabular}{@{}l@{ }c@{ }c@{}}
    \toprule
    \textbf{Model} & \textbf{Dial Avg RMSE} & \textbf{Norm Avg RMSE} \\
    \midrule
    ElasticNet                   & \textbf{2.93}  &  3.25 \\
    K Nearest Neighbors          & \textbf{3.16} &  3.24 \\
    Linear Regression            & \textbf{2.97}  &  3.32 \\
    Random Forest                & \textbf{2.98}  &  3.20 \\
    ERT   & \textbf{2.99}  &  3.23 \\
    \bottomrule
    \end{tabular}
    \caption{Average root mean square error (RMSE) per model for coordinate regression, using the original dialectal dataset and our normalized one. Regression is harder after removing superficial linguistic information, but not as much as classification.}
    \label{tab:regr_summary}
\end{table}

\subsection{Region Clustering}
Using \textbf{GreekBERT}, a monolingual encoder-only model for Standard Modern Greek \cite{Koutsikakis_2020}, we construct a dense representation for each proverb by averaging the embeddings of its tokens. We then average the representations of all proverbs for each region to create representations of the regions themselves, and perform clustering of the regions. As input, we use both the original corpus provided in our dataset, as well as the normalized one. No geolocation data is provided; only the text of the proverbs from each region.

\begin{figure}[t]
    \centering
    \small
    \begin{tabular}{c}
       \includegraphics[width=1\linewidth]{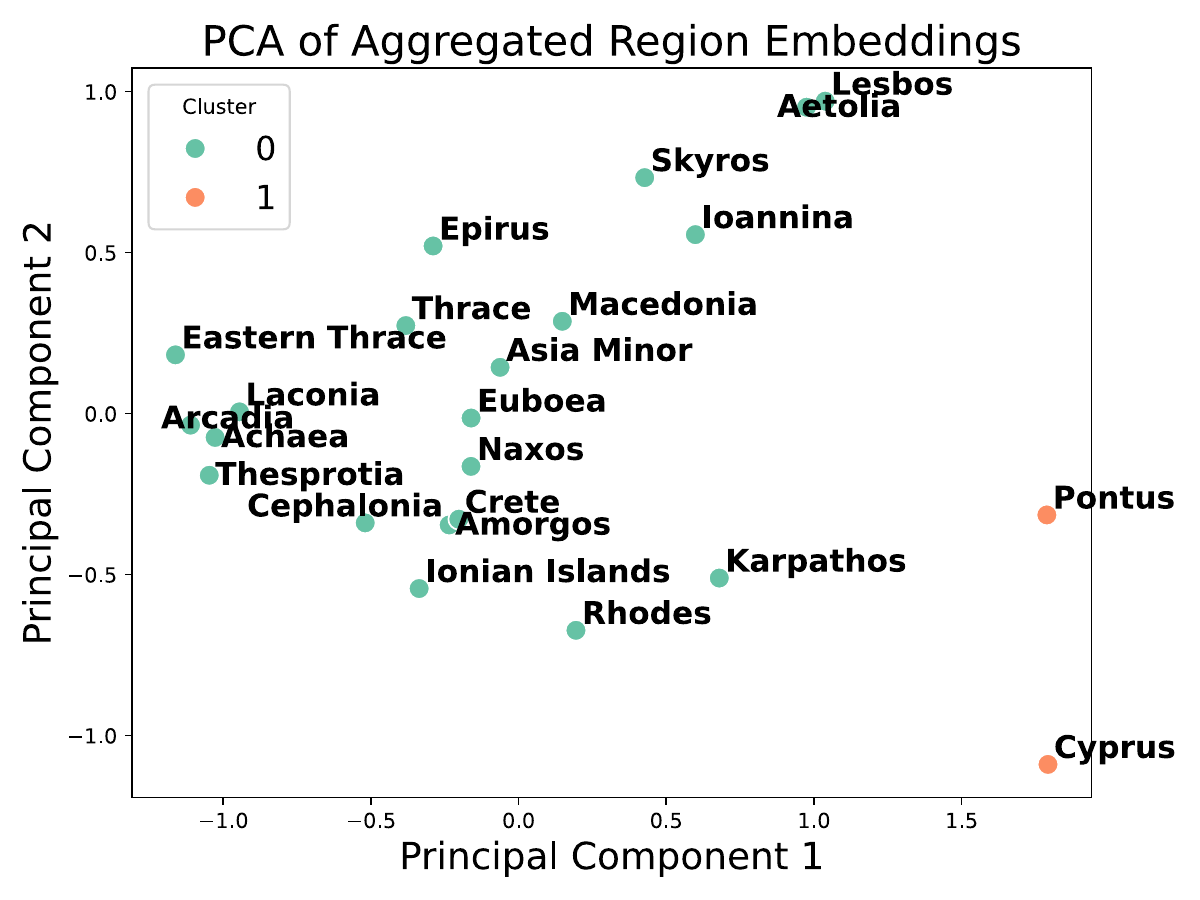} \\
        (a) Original Dialectal Data \\
       \includegraphics[width=1\linewidth]{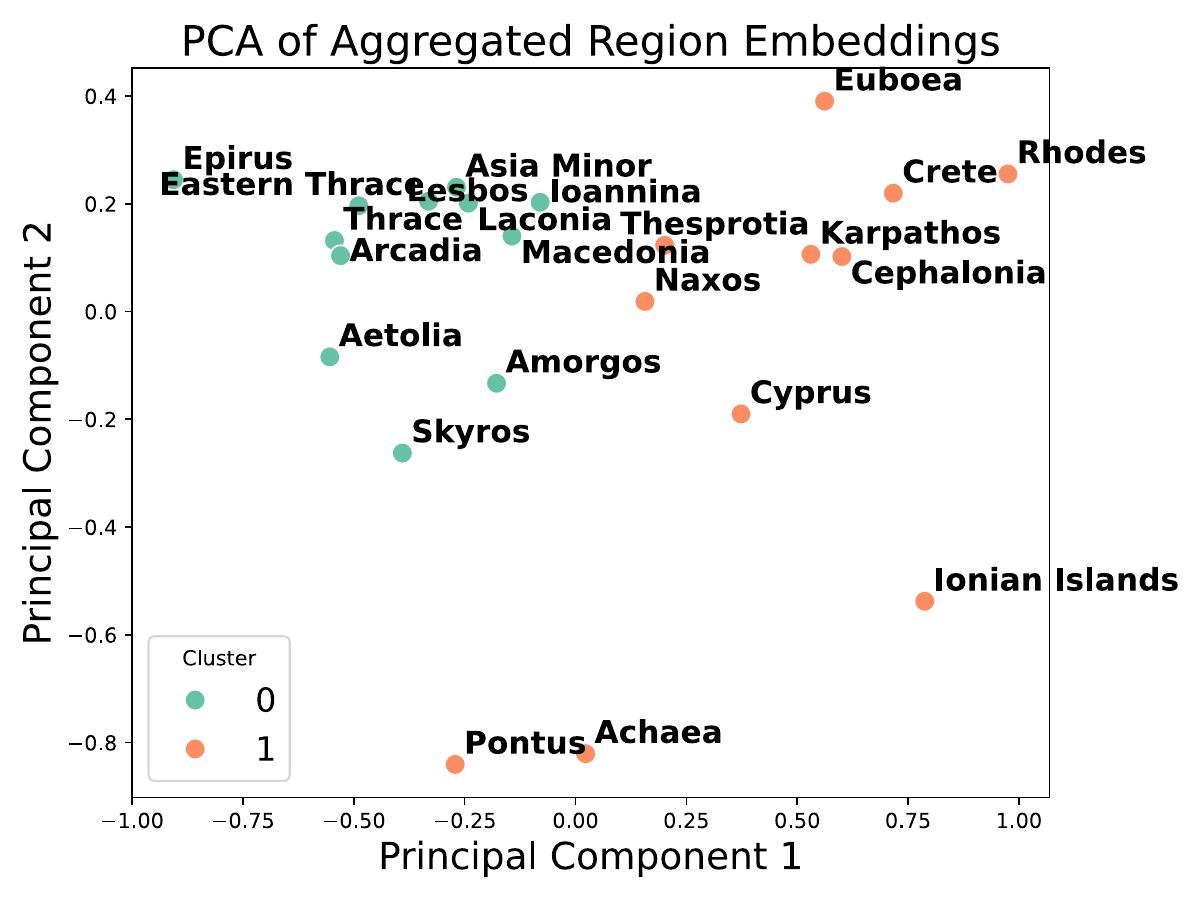} \\
        (b) Normalized Data 
    \end{tabular}
    \caption{$k$-means clustering with normalized data produces more reasonable clusters (full size in App.~\ref{app:kmeans}).}
    \label{fig:clustboth}
\end{figure}

Using K-means and the Silhouette method, we find the best results in both dataset versions are obtained for \(k=2\). The outputs of other algorithms, including of K-means for different values of K, can be found in Appendix~\ref{sec:clust}.
The output of the algorithm using the two versions of the data is shown in Figure~\ref{fig:clustboth}.
Based on these depictions, we consider that the results are far more meaningful when the data are normalized first. Using the original dialectal data, Pontus and Cyprus, two distant and unrelated regions, are put together in one cluster, and everything else is clustered together. With our normalized version, one cluster consists of islands and coastal regions, and the other of mainland ones. The few outliers, such as Skyros and Lesbos, are not random either. Despite being islands, they appear in the ``mainland'' cluster, but they are also the only islands in our dataset that have historically had significant connections with the Northern mainland. Overall, while there is no clearly discernible geographic information in the PCA plot produced using the dialectal data, the normalized one seems to have roughly put Western and Northern regions on the top and left, while Eastern and Southern ones are on the bottom and right. This implies that we can now uncover geographic information through the semantic similarity of proverbs.

\subsection{Cardinal direction driving words}
We also fine-tune GreekBERT (see Appendix~\ref{sec:hyper}) to predict geographic coordinates (as in \S\ref{ssec:geolocation}, achieving a mean absolute error of 1.59). To analyze which tokens guide this model towards each cardinal direction, we iterate over the dataset and mask each token in every proverb, averaging the change in the predicted coordinates, in a method similar to input erasure \cite{pavlopoulos-etal-2021-semeval}. We find meaningful results, such as the words for ``cold'' and ``winter'' being among the most influential ones in pushing the prediction to the North, which has a significantly colder climate.

\section{Discussion and Future Work}

Our experimental results show that our full setup outclasses all tested baselines in terms of both form normalization and meaning conservation, but also independently achieves performance similar to an ideal human expert (who would have achieved scores close to the 5-point mark). This, along with the results achieved in downstream tasks, indicates that our approach can be used in various contexts for dialectal NLU as an upstream method.

When it comes to the downstream experiments, we hypothesize that the difference in performance between the old and new ones has to do with the different methods of dialectal transcription used for each region. Even though they appear to offer very clear signals for recognizing each area specifically, they obfuscate existing dialectal and cultural similarities. Therefore, when using normalized data, it is impossible to pinpoint exactly the area where a specific proverb originates from, as they are often widely shared. Conversely, it is much easier to categorize the regions themselves, as by removing the layer of transcription, which previously created unrelated ``islands'' for each specific region, interregional parallels can be detected.

Our method adds little additional overhead, monetary or temporal, to the baseline of simply using an LLM for the task, as RBN can be executed within seconds for our entire dataset on a consumer CPU.

Based on feedback from our annotators, we notice that the main failure case is sentences containing dialectal vocabulary without a clear cognate in Standard Modern Greek. Since such rare vocabulary does not appear in any of the LLMs' training data with sufficient frequency so that its meaning can be learned, and morphological rules cannot address purely lexical divergence from the standard, the model is left to infer the meaning from the surrounding context.

\paragraph{Future Work} \ 
We believe that it would be worthwhile to create comprehensive dictionaries of dialectal terms which do not appear in the standard, especially for varieties that are overall relatively close linguistically to a higher-resource language, in cases where they do not already exist (as is the case for most Greek dialects).

Given that our results indicate that this is the main issue currently complicating automatic processing for these dialects, at least when it comes to their understanding, such a resource could be a crucial tool in finally extending coverage to many underserved linguistic communities.

\FloatBarrier
\section*{Limitations}
We acknowledge that since the evaluators do not have native knowledge of all Greek dialects, they may have missed some of the subtle meanings of the proverbs whose translations they were evaluating. The sentences are, however, mostly understandable by all Greek speakers, and much of the normalization consisted of conforming to standard spelling.

Moreover, as mentioned above, our method may produce sub-par results in cases where rare dialectal vocabulary with no cognate in the standard is used.

\section*{Ethics Statement}
The very nature of the dialect-to-standard normalization task means that at least some sociolinguistic signals will be erased from the input, which risks contributing to the global decline in linguistic diversity. We do not consider this to be a significant additional risk, as our method is intended for Natural Language Understanding specifically. LLMs cannot on their own, using current architectures, learn such low-resource varieties, and dialectal text given to them in the form of prompts is already being stripped of much of its form and meaning by the model's internal processing. Our method is not intended for Natural Language Generation, which is to say, it would not cause a model to produce normalized instead of dialectal text, but it would help it better understand the dialect, which research indicates is the desire of the majority of dialectal speakers.

We have obtained permission from all annotators to publish the data they produced in the context of this paper. The annotators were volunteers, and no monetary compensation was provided for their involvement.

The content of the Greek Proverb Atlas Dataset is available under a CC BY-NC-ND 4.0 license, in csv format. Its usage in this project is therefore consistent with its intended use. All models we use come with permissive licenses, at least when it comes to research.

\section*{Acknowledgments}
The authors are grateful for the feedback from the anonymous reviewers and area chairs.
This work has been partially supported by project MIS 5154714 of the National Recovery and Resilience Plan Greece 2.0 funded by the European Union under the NextGenerationEU Program. Antonios Anastasopoulos is also supported by the US National Science Foundation under awards IIS-2125466 and IIS-2327143.

\bibliography{custom}

\begin{thebibliography}{18}
\expandafter\ifx\csname natexlab\endcsname\relax\def\natexlab#1{#1}\fi

\bibitem[{Abdul-Mageed et~al.(2023)Abdul-Mageed, Elmadany, Zhang, Nagoudi, Bouamor, and Habash}]{arabic}
Muhammad Abdul-Mageed, AbdelRahim Elmadany, Chiyu Zhang, El~Moatez~Billah Nagoudi, Houda Bouamor, and Nizar Habash. 2023.
\newblock \href {https://doi.org/10.18653/v1/2023.arabicnlp-1.62} {{NADI} 2023: The fourth nuanced {A}rabic dialect identification shared task}.
\newblock In \emph{Proceedings of ArabicNLP 2023}, pages 600--613, Singapore (Hybrid). Association for Computational Linguistics.

\bibitem[{Blaschke et~al.(2024)Blaschke, Purschke, Schuetze, and Plank}]{blaschke-etal-2024-dialect}
Verena Blaschke, Christoph Purschke, Hinrich Schuetze, and Barbara Plank. 2024.
\newblock \href {https://doi.org/10.18653/v1/2024.acl-short.74} {What do dialect speakers want? a survey of attitudes towards language technology for {G}erman dialects}.
\newblock In \emph{Proceedings of the 62nd Annual Meeting of the Association for Computational Linguistics (Volume 2: Short Papers)}, pages 823--841, Bangkok, Thailand. Association for Computational Linguistics.

\bibitem[{Chakravarthi et~al.(2021)Chakravarthi, Mihaela, Ionescu, Jauhiainen, Jauhiainen, Lind{\'e}n, Ljube{\v{s}}i{\'c}, Partanen, Priyadharshini, Purschke, Rajagopal, Scherrer, and Zampieri}]{chakravarthi-etal-2021-findings-vardial}
Bharathi~Raja Chakravarthi, Gaman Mihaela, Radu~Tudor Ionescu, Heidi Jauhiainen, Tommi Jauhiainen, Krister Lind{\'e}n, Nikola Ljube{\v{s}}i{\'c}, Niko Partanen, Ruba Priyadharshini, Christoph Purschke, Eswari Rajagopal, Yves Scherrer, and Marcos Zampieri. 2021.
\newblock \href {https://aclanthology.org/2021.vardial-1.1/} {Findings of the {V}ar{D}ial evaluation campaign 2021}.
\newblock In \emph{Proceedings of the Eighth Workshop on NLP for Similar Languages, Varieties and Dialects}, pages 1--11, Kiyv, Ukraine. Association for Computational Linguistics.

\bibitem[{Grattafiori et~al.(2024)Grattafiori, Dubey, Jauhri, Pandey, Kadian, Al-Dahle, Letman, Mathur, Schelten, Vaughan, Yang, Fan, Goyal, Hartshorn, Yang, Mitra, Sravankumar, Korenev, Hinsvark, Rao, Zhang, Rodriguez, Gregerson, Spataru, Roziere, Biron, Tang, Chern, Caucheteux, Nayak, Bi, Marra, McConnell, Keller, Touret, Wu, Wong, Ferrer, Nikolaidis, Allonsius, Song, Pintz, Livshits, Wyatt, Esiobu, Choudhary, Mahajan, Garcia-Olano, Perino, Hupkes, Lakomkin, AlBadawy, Lobanova, Dinan, Smith, Radenovic, Guzmán, Zhang, Synnaeve, Lee, Anderson, Thattai, Nail, Mialon, Pang, Cucurell, Nguyen, Korevaar, Xu, Touvron, Zarov, Ibarra, Kloumann, Misra, Evtimov, Zhang, Copet, Lee, Geffert, Vranes, Park, Mahadeokar, Shah, van~der Linde, Billock, Hong, Lee, Fu, Chi, Huang, Liu, Wang, Yu, Bitton, Spisak, Park, Rocca, Johnstun, Saxe, Jia, Alwala, Prasad, Upasani, Plawiak, Li, Heafield, Stone, El-Arini, Iyer, Malik, Chiu, Bhalla, Lakhotia, Rantala-Yeary, van~der Maaten, Chen, Tan, Jenkins, Martin, Madaan, Malo, Blecher,
  Landzaat, de~Oliveira, Muzzi, Pasupuleti, Singh, Paluri, Kardas, Tsimpoukelli, Oldham, Rita, Pavlova, Kambadur, Lewis, Si, Singh, Hassan, Goyal, Torabi, Bashlykov, Bogoychev, Chatterji, Zhang, Duchenne, Çelebi, Alrassy, Zhang, Li, Vasic, Weng, Bhargava, Dubal, Krishnan, Koura, Xu, He, Dong, Srinivasan, Ganapathy, Calderer, Cabral, Stojnic, Raileanu, Maheswari, Girdhar, Patel, Sauvestre, Polidoro, Sumbaly, Taylor, Silva, Hou, Wang, Hosseini, Chennabasappa, Singh, Bell, Kim, Edunov, Nie, Narang, Raparthy, Shen, Wan, Bhosale, Zhang, Vandenhende, Batra, Whitman, Sootla, Collot, Gururangan, Borodinsky, Herman, Fowler, Sheasha, Georgiou, Scialom, Speckbacher, Mihaylov, Xiao, Karn, Goswami, Gupta, Ramanathan, Kerkez, Gonguet, Do, Vogeti, Albiero, Petrovic, Chu, Xiong, Fu, Meers, Martinet, Wang, Wang, Tan, Xia, Xie, Jia, Wang, Goldschlag, Gaur, Babaei, Wen, Song, Zhang, Li, Mao, Coudert, Yan, Chen, Papakipos, Singh, Srivastava, Jain, Kelsey, Shajnfeld, Gangidi, Victoria, Goldstand, Menon, Sharma, Boesenberg,
  Baevski, Feinstein, Kallet, Sangani, Teo, Yunus, Lupu, Alvarado, Caples, Gu, Ho, Poulton, Ryan, Ramchandani, Dong, Franco, Goyal, Saraf, Chowdhury, Gabriel, Bharambe, Eisenman, Yazdan, James, Maurer, Leonhardi, Huang, Loyd, Paola, Paranjape, Liu, Wu, Ni, Hancock, Wasti, Spence, Stojkovic, Gamido, Montalvo, Parker, Burton, Mejia, Liu, Wang, Kim, Zhou, Hu, Chu, Cai, Tindal, Feichtenhofer, Gao, Civin, Beaty, Kreymer, Li, Adkins, Xu, Testuggine, David, Parikh, Liskovich, Foss, Wang, Le, Holland, Dowling, Jamil, Montgomery, Presani, Hahn, Wood, Le, Brinkman, Arcaute, Dunbar, Smothers, Sun, Kreuk, Tian, Kokkinos, Ozgenel, Caggioni, Kanayet, Seide, Florez, Schwarz, Badeer, Swee, Halpern, Herman, Sizov, Guangyi, Zhang, Lakshminarayanan, Inan, Shojanazeri, Zou, Wang, Zha, Habeeb, Rudolph, Suk, Aspegren, Goldman, Zhan, Damlaj, Molybog, Tufanov, Leontiadis, Veliche, Gat, Weissman, Geboski, Kohli, Lam, Asher, Gaya, Marcus, Tang, Chan, Zhen, Reizenstein, Teboul, Zhong, Jin, Yang, Cummings, Carvill, Shepard, McPhie,
  Torres, Ginsburg, Wang, Wu, U, Saxena, Khandelwal, Zand, Matosich, Veeraraghavan, Michelena, Li, Jagadeesh, Huang, Chawla, Huang, Chen, Garg, A, Silva, Bell, Zhang, Guo, Yu, Moshkovich, Wehrstedt, Khabsa, Avalani, Bhatt, Mankus, Hasson, Lennie, Reso, Groshev, Naumov, Lathi, Keneally, Liu, Seltzer, Valko, Restrepo, Patel, Vyatskov, Samvelyan, Clark, Macey, Wang, Hermoso, Metanat, Rastegari, Bansal, Santhanam, Parks, White, Bawa, Singhal, Egebo, Usunier, Mehta, Laptev, Dong, Cheng, Chernoguz, Hart, Salpekar, Kalinli, Kent, Parekh, Saab, Balaji, Rittner, Bontrager, Roux, Dollar, Zvyagina, Ratanchandani, Yuvraj, Liang, Alao, Rodriguez, Ayub, Murthy, Nayani, Mitra, Parthasarathy, Li, Hogan, Battey, Wang, Howes, Rinott, Mehta, Siby, Bondu, Datta, Chugh, Hunt, Dhillon, Sidorov, Pan, Mahajan, Verma, Yamamoto, Ramaswamy, Lindsay, Lindsay, Feng, Lin, Zha, Patil, Shankar, Zhang, Zhang, Wang, Agarwal, Sajuyigbe, Chintala, Max, Chen, Kehoe, Satterfield, Govindaprasad, Gupta, Deng, Cho, Virk, Subramanian, Choudhury,
  Goldman, Remez, Glaser, Best, Koehler, Robinson, Li, Zhang, Matthews, Chou, Shaked, Vontimitta, Ajayi, Montanez, Mohan, Kumar, Mangla, Ionescu, Poenaru, Mihailescu, Ivanov, Li, Wang, Jiang, Bouaziz, Constable, Tang, Wu, Wang, Wu, Gao, Kleinman, Chen, Hu, Jia, Qi, Li, Zhang, Zhang, Adi, Nam, Yu, Wang, Zhao, Hao, Qian, Li, He, Rait, DeVito, Rosnbrick, Wen, Yang, Zhao, and Ma}]{grattafiori2024llama3herdmodels}
Aaron Grattafiori, Abhimanyu Dubey, Abhinav Jauhri, Abhinav Pandey, Abhishek Kadian, Ahmad Al-Dahle, Aiesha Letman, Akhil Mathur, Alan Schelten, Alex Vaughan, Amy Yang, Angela Fan, Anirudh Goyal, Anthony Hartshorn, Aobo Yang, Archi Mitra, Archie Sravankumar, Artem Korenev, Arthur Hinsvark, Arun Rao, Aston Zhang, Aurelien Rodriguez, Austen Gregerson, Ava Spataru, Baptiste Roziere, Bethany Biron, Binh Tang, Bobbie Chern, Charlotte Caucheteux, Chaya Nayak, Chloe Bi, Chris Marra, Chris McConnell, Christian Keller, Christophe Touret, Chunyang Wu, Corinne Wong, Cristian~Canton Ferrer, Cyrus Nikolaidis, Damien Allonsius, Daniel Song, Danielle Pintz, Danny Livshits, Danny Wyatt, David Esiobu, Dhruv Choudhary, Dhruv Mahajan, Diego Garcia-Olano, Diego Perino, Dieuwke Hupkes, Egor Lakomkin, Ehab AlBadawy, Elina Lobanova, Emily Dinan, Eric~Michael Smith, Filip Radenovic, Francisco Guzmán, Frank Zhang, Gabriel Synnaeve, Gabrielle Lee, Georgia~Lewis Anderson, Govind Thattai, Graeme Nail, Gregoire Mialon, Guan Pang,
  Guillem Cucurell, Hailey Nguyen, Hannah Korevaar, Hu~Xu, Hugo Touvron, Iliyan Zarov, Imanol~Arrieta Ibarra, Isabel Kloumann, Ishan Misra, Ivan Evtimov, Jack Zhang, Jade Copet, Jaewon Lee, Jan Geffert, Jana Vranes, Jason Park, Jay Mahadeokar, Jeet Shah, Jelmer van~der Linde, Jennifer Billock, Jenny Hong, Jenya Lee, Jeremy Fu, Jianfeng Chi, Jianyu Huang, Jiawen Liu, Jie Wang, Jiecao Yu, Joanna Bitton, Joe Spisak, Jongsoo Park, Joseph Rocca, Joshua Johnstun, Joshua Saxe, Junteng Jia, Kalyan~Vasuden Alwala, Karthik Prasad, Kartikeya Upasani, Kate Plawiak, Ke~Li, Kenneth Heafield, Kevin Stone, Khalid El-Arini, Krithika Iyer, Kshitiz Malik, Kuenley Chiu, Kunal Bhalla, Kushal Lakhotia, Lauren Rantala-Yeary, Laurens van~der Maaten, Lawrence Chen, Liang Tan, Liz Jenkins, Louis Martin, Lovish Madaan, Lubo Malo, Lukas Blecher, Lukas Landzaat, Luke de~Oliveira, Madeline Muzzi, Mahesh Pasupuleti, Mannat Singh, Manohar Paluri, Marcin Kardas, Maria Tsimpoukelli, Mathew Oldham, Mathieu Rita, Maya Pavlova, Melanie Kambadur,
  Mike Lewis, Min Si, Mitesh~Kumar Singh, Mona Hassan, Naman Goyal, Narjes Torabi, Nikolay Bashlykov, Nikolay Bogoychev, Niladri Chatterji, Ning Zhang, Olivier Duchenne, Onur Çelebi, Patrick Alrassy, Pengchuan Zhang, Pengwei Li, Petar Vasic, Peter Weng, Prajjwal Bhargava, Pratik Dubal, Praveen Krishnan, Punit~Singh Koura, Puxin Xu, Qing He, Qingxiao Dong, Ragavan Srinivasan, Raj Ganapathy, Ramon Calderer, Ricardo~Silveira Cabral, Robert Stojnic, Roberta Raileanu, Rohan Maheswari, Rohit Girdhar, Rohit Patel, Romain Sauvestre, Ronnie Polidoro, Roshan Sumbaly, Ross Taylor, Ruan Silva, Rui Hou, Rui Wang, Saghar Hosseini, Sahana Chennabasappa, Sanjay Singh, Sean Bell, Seohyun~Sonia Kim, Sergey Edunov, Shaoliang Nie, Sharan Narang, Sharath Raparthy, Sheng Shen, Shengye Wan, Shruti Bhosale, Shun Zhang, Simon Vandenhende, Soumya Batra, Spencer Whitman, Sten Sootla, Stephane Collot, Suchin Gururangan, Sydney Borodinsky, Tamar Herman, Tara Fowler, Tarek Sheasha, Thomas Georgiou, Thomas Scialom, Tobias Speckbacher,
  Todor Mihaylov, Tong Xiao, Ujjwal Karn, Vedanuj Goswami, Vibhor Gupta, Vignesh Ramanathan, Viktor Kerkez, Vincent Gonguet, Virginie Do, Vish Vogeti, Vítor Albiero, Vladan Petrovic, Weiwei Chu, Wenhan Xiong, Wenyin Fu, Whitney Meers, Xavier Martinet, Xiaodong Wang, Xiaofang Wang, Xiaoqing~Ellen Tan, Xide Xia, Xinfeng Xie, Xuchao Jia, Xuewei Wang, Yaelle Goldschlag, Yashesh Gaur, Yasmine Babaei, Yi~Wen, Yiwen Song, Yuchen Zhang, Yue Li, Yuning Mao, Zacharie~Delpierre Coudert, Zheng Yan, Zhengxing Chen, Zoe Papakipos, Aaditya Singh, Aayushi Srivastava, Abha Jain, Adam Kelsey, Adam Shajnfeld, Adithya Gangidi, Adolfo Victoria, Ahuva Goldstand, Ajay Menon, Ajay Sharma, Alex Boesenberg, Alexei Baevski, Allie Feinstein, Amanda Kallet, Amit Sangani, Amos Teo, Anam Yunus, Andrei Lupu, Andres Alvarado, Andrew Caples, Andrew Gu, Andrew Ho, Andrew Poulton, Andrew Ryan, Ankit Ramchandani, Annie Dong, Annie Franco, Anuj Goyal, Aparajita Saraf, Arkabandhu Chowdhury, Ashley Gabriel, Ashwin Bharambe, Assaf Eisenman, Azadeh
  Yazdan, Beau James, Ben Maurer, Benjamin Leonhardi, Bernie Huang, Beth Loyd, Beto~De Paola, Bhargavi Paranjape, Bing Liu, Bo~Wu, Boyu Ni, Braden Hancock, Bram Wasti, Brandon Spence, Brani Stojkovic, Brian Gamido, Britt Montalvo, Carl Parker, Carly Burton, Catalina Mejia, Ce~Liu, Changhan Wang, Changkyu Kim, Chao Zhou, Chester Hu, Ching-Hsiang Chu, Chris Cai, Chris Tindal, Christoph Feichtenhofer, Cynthia Gao, Damon Civin, Dana Beaty, Daniel Kreymer, Daniel Li, David Adkins, David Xu, Davide Testuggine, Delia David, Devi Parikh, Diana Liskovich, Didem Foss, Dingkang Wang, Duc Le, Dustin Holland, Edward Dowling, Eissa Jamil, Elaine Montgomery, Eleonora Presani, Emily Hahn, Emily Wood, Eric-Tuan Le, Erik Brinkman, Esteban Arcaute, Evan Dunbar, Evan Smothers, Fei Sun, Felix Kreuk, Feng Tian, Filippos Kokkinos, Firat Ozgenel, Francesco Caggioni, Frank Kanayet, Frank Seide, Gabriela~Medina Florez, Gabriella Schwarz, Gada Badeer, Georgia Swee, Gil Halpern, Grant Herman, Grigory Sizov, Guangyi, Zhang, Guna
  Lakshminarayanan, Hakan Inan, Hamid Shojanazeri, Han Zou, Hannah Wang, Hanwen Zha, Haroun Habeeb, Harrison Rudolph, Helen Suk, Henry Aspegren, Hunter Goldman, Hongyuan Zhan, Ibrahim Damlaj, Igor Molybog, Igor Tufanov, Ilias Leontiadis, Irina-Elena Veliche, Itai Gat, Jake Weissman, James Geboski, James Kohli, Janice Lam, Japhet Asher, Jean-Baptiste Gaya, Jeff Marcus, Jeff Tang, Jennifer Chan, Jenny Zhen, Jeremy Reizenstein, Jeremy Teboul, Jessica Zhong, Jian Jin, Jingyi Yang, Joe Cummings, Jon Carvill, Jon Shepard, Jonathan McPhie, Jonathan Torres, Josh Ginsburg, Junjie Wang, Kai Wu, Kam~Hou U, Karan Saxena, Kartikay Khandelwal, Katayoun Zand, Kathy Matosich, Kaushik Veeraraghavan, Kelly Michelena, Keqian Li, Kiran Jagadeesh, Kun Huang, Kunal Chawla, Kyle Huang, Lailin Chen, Lakshya Garg, Lavender A, Leandro Silva, Lee Bell, Lei Zhang, Liangpeng Guo, Licheng Yu, Liron Moshkovich, Luca Wehrstedt, Madian Khabsa, Manav Avalani, Manish Bhatt, Martynas Mankus, Matan Hasson, Matthew Lennie, Matthias Reso, Maxim
  Groshev, Maxim Naumov, Maya Lathi, Meghan Keneally, Miao Liu, Michael~L. Seltzer, Michal Valko, Michelle Restrepo, Mihir Patel, Mik Vyatskov, Mikayel Samvelyan, Mike Clark, Mike Macey, Mike Wang, Miquel~Jubert Hermoso, Mo~Metanat, Mohammad Rastegari, Munish Bansal, Nandhini Santhanam, Natascha Parks, Natasha White, Navyata Bawa, Nayan Singhal, Nick Egebo, Nicolas Usunier, Nikhil Mehta, Nikolay~Pavlovich Laptev, Ning Dong, Norman Cheng, Oleg Chernoguz, Olivia Hart, Omkar Salpekar, Ozlem Kalinli, Parkin Kent, Parth Parekh, Paul Saab, Pavan Balaji, Pedro Rittner, Philip Bontrager, Pierre Roux, Piotr Dollar, Polina Zvyagina, Prashant Ratanchandani, Pritish Yuvraj, Qian Liang, Rachad Alao, Rachel Rodriguez, Rafi Ayub, Raghotham Murthy, Raghu Nayani, Rahul Mitra, Rangaprabhu Parthasarathy, Raymond Li, Rebekkah Hogan, Robin Battey, Rocky Wang, Russ Howes, Ruty Rinott, Sachin Mehta, Sachin Siby, Sai~Jayesh Bondu, Samyak Datta, Sara Chugh, Sara Hunt, Sargun Dhillon, Sasha Sidorov, Satadru Pan, Saurabh Mahajan,
  Saurabh Verma, Seiji Yamamoto, Sharadh Ramaswamy, Shaun Lindsay, Shaun Lindsay, Sheng Feng, Shenghao Lin, Shengxin~Cindy Zha, Shishir Patil, Shiva Shankar, Shuqiang Zhang, Shuqiang Zhang, Sinong Wang, Sneha Agarwal, Soji Sajuyigbe, Soumith Chintala, Stephanie Max, Stephen Chen, Steve Kehoe, Steve Satterfield, Sudarshan Govindaprasad, Sumit Gupta, Summer Deng, Sungmin Cho, Sunny Virk, Suraj Subramanian, Sy~Choudhury, Sydney Goldman, Tal Remez, Tamar Glaser, Tamara Best, Thilo Koehler, Thomas Robinson, Tianhe Li, Tianjun Zhang, Tim Matthews, Timothy Chou, Tzook Shaked, Varun Vontimitta, Victoria Ajayi, Victoria Montanez, Vijai Mohan, Vinay~Satish Kumar, Vishal Mangla, Vlad Ionescu, Vlad Poenaru, Vlad~Tiberiu Mihailescu, Vladimir Ivanov, Wei Li, Wenchen Wang, Wenwen Jiang, Wes Bouaziz, Will Constable, Xiaocheng Tang, Xiaojian Wu, Xiaolan Wang, Xilun Wu, Xinbo Gao, Yaniv Kleinman, Yanjun Chen, Ye~Hu, Ye~Jia, Ye~Qi, Yenda Li, Yilin Zhang, Ying Zhang, Yossi Adi, Youngjin Nam, Yu, Wang, Yu~Zhao, Yuchen Hao, Yundi
  Qian, Yunlu Li, Yuzi He, Zach Rait, Zachary DeVito, Zef Rosnbrick, Zhaoduo Wen, Zhenyu Yang, Zhiwei Zhao, and Zhiyu Ma. 2024.
\newblock \href {http://arxiv.org/abs/2407.21783} {The llama 3 herd of models}.

\bibitem[{Han et~al.(2016)Han, Rahimi, Derczynski, and Baldwin}]{han-etal-2016-twitter}
Bo~Han, Afshin Rahimi, Leon Derczynski, and Timothy Baldwin. 2016.
\newblock \href {https://aclanthology.org/W16-3928/} {{T}witter geolocation prediction shared task of the 2016 workshop on noisy user-generated text}.
\newblock In \emph{Proceedings of the 2nd Workshop on Noisy User-generated Text ({WNUT})}, pages 213--217, Osaka, Japan. The COLING 2016 Organizing Committee.

\bibitem[{Hovy and Purschke(2018)}]{hovy-purschke-2018-capturing}
Dirk Hovy and Christoph Purschke. 2018.
\newblock \href {https://doi.org/10.18653/v1/D18-1469} {Capturing regional variation with distributed place representations and geographic retrofitting}.
\newblock In \emph{Proceedings of the 2018 Conference on Empirical Methods in Natural Language Processing}, pages 4383--4394, Brussels, Belgium. Association for Computational Linguistics.

\bibitem[{Ibn~Alam and Anastasopoulos(2025)}]{alam-anastasopoulos-2025-normalization}
Md~Mahfuz Ibn~Alam and Antonios Anastasopoulos. 2025.
\newblock \href {#} {Large language models as a normalizer for transliteration and dialectal translation}.
\newblock In \emph{Proceedings of the Twelfth Workshop on NLP for Similar Languages, Varieties, and Dialects (VarDial 2025)}, Abu Dhabi, UAE. Association for Computational Linguistics.

\bibitem[{Joshi et~al.(2025)Joshi, Dabre, Kanojia, Li, Zhan, Haffari, and Dippold}]{10.1145/3712060}
Aditya Joshi, Raj Dabre, Diptesh Kanojia, Zhuang Li, Haolan Zhan, Gholamreza Haffari, and Doris Dippold. 2025.
\newblock \href {https://doi.org/10.1145/3712060} {Natural language processing for dialects of a language: A survey}.
\newblock \emph{ACM Comput. Surv.}, 57(6).

\bibitem[{Koo and Li(2016)}]{icc}
Terry~K Koo and Mae~Y Li. 2016.
\newblock A guideline of selecting and reporting intraclass correlation coefficients for reliability research.
\newblock \emph{J. Chiropr. Med.}, 15(2):155--163.

\bibitem[{Koutsikakis et~al.(2020)Koutsikakis, Chalkidis, Malakasiotis, and Androutsopoulos}]{Koutsikakis_2020}
John Koutsikakis, Ilias Chalkidis, Prodromos Malakasiotis, and Ion Androutsopoulos. 2020.
\newblock \href {https://doi.org/10.1145/3411408.3411440} {Greek-bert: The greeks visiting sesame street}.
\newblock In \emph{11th Hellenic Conference on Artificial Intelligence}, SETN 2020, page 110–117. ACM.

\bibitem[{Kuparinen et~al.(2023)Kuparinen, Mileti{\'c}, and Scherrer}]{general}
Olli Kuparinen, Aleksandra Mileti{\'c}, and Yves Scherrer. 2023.
\newblock \href {https://doi.org/10.18653/v1/2023.findings-emnlp.923} {Dialect-to-standard normalization: A large-scale multilingual evaluation}.
\newblock In \emph{Findings of the Association for Computational Linguistics: EMNLP 2023}, pages 13814--13828, Singapore. Association for Computational Linguistics.

\bibitem[{OpenAI et~al.(2024)OpenAI, :, Hurst, Lerer, Goucher, Perelman, Ramesh, Clark, Ostrow, Welihinda, Hayes, Radford, Mądry, Baker-Whitcomb, Beutel, Borzunov, Carney, Chow, Kirillov, Nichol, Paino, Renzin, Passos, Kirillov, Christakis, Conneau, Kamali, Jabri, Moyer, Tam, Crookes, Tootoochian, Tootoonchian, Kumar, Vallone, Karpathy, Braunstein, Cann, Codispoti, Galu, Kondrich, Tulloch, Mishchenko, Baek, Jiang, Pelisse, Woodford, Gosalia, Dhar, Pantuliano, Nayak, Oliver, Zoph, Ghorbani, Leimberger, Rossen, Sokolowsky, Wang, Zweig, Hoover, Samic, McGrew, Spero, Giertler, Cheng, Lightcap, Walkin, Quinn, Guarraci, Hsu, Kellogg, Eastman, Lugaresi, Wainwright, Bassin, Hudson, Chu, Nelson, Li, Shern, Conger, Barette, Voss, Ding, Lu, Zhang, Beaumont, Hallacy, Koch, Gibson, Kim, Choi, McLeavey, Hesse, Fischer, Winter, Czarnecki, Jarvis, Wei, Koumouzelis, Sherburn, Kappler, Levin, Levy, Carr, Farhi, Mely, Robinson, Sasaki, Jin, Valladares, Tsipras, Li, Nguyen, Findlay, Oiwoh, Wong, Asdar, Proehl, Yang, Antonow,
  Kramer, Peterson, Sigler, Wallace, Brevdo, Mays, Khorasani, Such, Raso, Zhang, von Lohmann, Sulit, Goh, Oden, Salmon, Starace, Brockman, Salman, Bao, Hu, Wong, Wang, Schmidt, Whitney, Jun, Kirchner, de~Oliveira~Pinto, Ren, Chang, Chung, Kivlichan, O'Connell, O'Connell, Osband, Silber, Sohl, Okuyucu, Lan, Kostrikov, Sutskever, Kanitscheider, Gulrajani, Coxon, Menick, Pachocki, Aung, Betker, Crooks, Lennon, Kiros, Leike, Park, Kwon, Phang, Teplitz, Wei, Wolfe, Chen, Harris, Varavva, Lee, Shieh, Lin, Yu, Weng, Tang, Yu, Jang, Candela, Beutler, Landers, Parish, Heidecke, Schulman, Lachman, McKay, Uesato, Ward, Kim, Huizinga, Sitkin, Kraaijeveld, Gross, Kaplan, Snyder, Achiam, Jiao, Lee, Zhuang, Harriman, Fricke, Hayashi, Singhal, Shi, Karthik, Wood, Rimbach, Hsu, Nguyen, Gu-Lemberg, Button, Liu, Howe, Muthukumar, Luther, Ahmad, Kai, Itow, Workman, Pathak, Chen, Jing, Guy, Fedus, Zhou, Mamitsuka, Weng, McCallum, Held, Ouyang, Feuvrier, Zhang, Kondraciuk, Kaiser, Hewitt, Metz, Doshi, Aflak, Simens, Boyd,
  Thompson, Dukhan, Chen, Gray, Hudnall, Zhang, Aljubeh, Litwin, Zeng, Johnson, Shetty, Gupta, Shah, Yatbaz, Yang, Zhong, Glaese, Chen, Janner, Lampe, Petrov, Wu, Wang, Fradin, Pokrass, Castro, de~Castro, Pavlov, Brundage, Wang, Khan, Murati, Bavarian, Lin, Yesildal, Soto, Gimelshein, Cone, Staudacher, Summers, LaFontaine, Chowdhury, Ryder, Stathas, Turley, Tezak, Felix, Kudige, Keskar, Deutsch, Bundick, Puckett, Nachum, Okelola, Boiko, Murk, Jaffe, Watkins, Godement, Campbell-Moore, Chao, McMillan, Belov, Su, Bak, Bakkum, Deng, Dolan, Hoeschele, Welinder, Tillet, Pronin, Tillet, Dhariwal, Yuan, Dias, Lim, Arora, Troll, Lin, Lopes, Puri, Miyara, Leike, Gaubert, Zamani, Wang, Donnelly, Honsby, Smith, Sahai, Ramchandani, Huet, Carmichael, Zellers, Chen, Chen, Nigmatullin, Cheu, Jain, Altman, Schoenholz, Toizer, Miserendino, Agarwal, Culver, Ethersmith, Gray, Grove, Metzger, Hermani, Jain, Zhao, Wu, Jomoto, Wu, Shuaiqi, Xia, Phene, Papay, Narayanan, Coffey, Lee, Hall, Balaji, Broda, Stramer, Xu, Gogineni,
  Christianson, Sanders, Patwardhan, Cunninghman, Degry, Dimson, Raoux, Shadwell, Zheng, Underwood, Markov, Sherbakov, Rubin, Stasi, Kaftan, Heywood, Peterson, Walters, Eloundou, Qi, Moeller, Monaco, Kuo, Fomenko, Chang, Zheng, Zhou, Manassra, Sheu, Zaremba, Patil, Qian, Kim, Cheng, Zhang, He, Zhang, Jin, Dai, and Malkov}]{openai2024gpt4ocard}
OpenAI, :, Aaron Hurst, Adam Lerer, Adam~P. Goucher, Adam Perelman, Aditya Ramesh, Aidan Clark, AJ~Ostrow, Akila Welihinda, Alan Hayes, Alec Radford, Aleksander Mądry, Alex Baker-Whitcomb, Alex Beutel, Alex Borzunov, Alex Carney, Alex Chow, Alex Kirillov, Alex Nichol, Alex Paino, Alex Renzin, Alex~Tachard Passos, Alexander Kirillov, Alexi Christakis, Alexis Conneau, Ali Kamali, Allan Jabri, Allison Moyer, Allison Tam, Amadou Crookes, Amin Tootoochian, Amin Tootoonchian, Ananya Kumar, Andrea Vallone, Andrej Karpathy, Andrew Braunstein, Andrew Cann, Andrew Codispoti, Andrew Galu, Andrew Kondrich, Andrew Tulloch, Andrey Mishchenko, Angela Baek, Angela Jiang, Antoine Pelisse, Antonia Woodford, Anuj Gosalia, Arka Dhar, Ashley Pantuliano, Avi Nayak, Avital Oliver, Barret Zoph, Behrooz Ghorbani, Ben Leimberger, Ben Rossen, Ben Sokolowsky, Ben Wang, Benjamin Zweig, Beth Hoover, Blake Samic, Bob McGrew, Bobby Spero, Bogo Giertler, Bowen Cheng, Brad Lightcap, Brandon Walkin, Brendan Quinn, Brian Guarraci, Brian Hsu,
  Bright Kellogg, Brydon Eastman, Camillo Lugaresi, Carroll Wainwright, Cary Bassin, Cary Hudson, Casey Chu, Chad Nelson, Chak Li, Chan~Jun Shern, Channing Conger, Charlotte Barette, Chelsea Voss, Chen Ding, Cheng Lu, Chong Zhang, Chris Beaumont, Chris Hallacy, Chris Koch, Christian Gibson, Christina Kim, Christine Choi, Christine McLeavey, Christopher Hesse, Claudia Fischer, Clemens Winter, Coley Czarnecki, Colin Jarvis, Colin Wei, Constantin Koumouzelis, Dane Sherburn, Daniel Kappler, Daniel Levin, Daniel Levy, David Carr, David Farhi, David Mely, David Robinson, David Sasaki, Denny Jin, Dev Valladares, Dimitris Tsipras, Doug Li, Duc~Phong Nguyen, Duncan Findlay, Edede Oiwoh, Edmund Wong, Ehsan Asdar, Elizabeth Proehl, Elizabeth Yang, Eric Antonow, Eric Kramer, Eric Peterson, Eric Sigler, Eric Wallace, Eugene Brevdo, Evan Mays, Farzad Khorasani, Felipe~Petroski Such, Filippo Raso, Francis Zhang, Fred von Lohmann, Freddie Sulit, Gabriel Goh, Gene Oden, Geoff Salmon, Giulio Starace, Greg Brockman, Hadi
  Salman, Haiming Bao, Haitang Hu, Hannah Wong, Haoyu Wang, Heather Schmidt, Heather Whitney, Heewoo Jun, Hendrik Kirchner, Henrique~Ponde de~Oliveira~Pinto, Hongyu Ren, Huiwen Chang, Hyung~Won Chung, Ian Kivlichan, Ian O'Connell, Ian O'Connell, Ian Osband, Ian Silber, Ian Sohl, Ibrahim Okuyucu, Ikai Lan, Ilya Kostrikov, Ilya Sutskever, Ingmar Kanitscheider, Ishaan Gulrajani, Jacob Coxon, Jacob Menick, Jakub Pachocki, James Aung, James Betker, James Crooks, James Lennon, Jamie Kiros, Jan Leike, Jane Park, Jason Kwon, Jason Phang, Jason Teplitz, Jason Wei, Jason Wolfe, Jay Chen, Jeff Harris, Jenia Varavva, Jessica~Gan Lee, Jessica Shieh, Ji~Lin, Jiahui Yu, Jiayi Weng, Jie Tang, Jieqi Yu, Joanne Jang, Joaquin~Quinonero Candela, Joe Beutler, Joe Landers, Joel Parish, Johannes Heidecke, John Schulman, Jonathan Lachman, Jonathan McKay, Jonathan Uesato, Jonathan Ward, Jong~Wook Kim, Joost Huizinga, Jordan Sitkin, Jos Kraaijeveld, Josh Gross, Josh Kaplan, Josh Snyder, Joshua Achiam, Joy Jiao, Joyce Lee, Juntang
  Zhuang, Justyn Harriman, Kai Fricke, Kai Hayashi, Karan Singhal, Katy Shi, Kavin Karthik, Kayla Wood, Kendra Rimbach, Kenny Hsu, Kenny Nguyen, Keren Gu-Lemberg, Kevin Button, Kevin Liu, Kiel Howe, Krithika Muthukumar, Kyle Luther, Lama Ahmad, Larry Kai, Lauren Itow, Lauren Workman, Leher Pathak, Leo Chen, Li~Jing, Lia Guy, Liam Fedus, Liang Zhou, Lien Mamitsuka, Lilian Weng, Lindsay McCallum, Lindsey Held, Long Ouyang, Louis Feuvrier, Lu~Zhang, Lukas Kondraciuk, Lukasz Kaiser, Luke Hewitt, Luke Metz, Lyric Doshi, Mada Aflak, Maddie Simens, Madelaine Boyd, Madeleine Thompson, Marat Dukhan, Mark Chen, Mark Gray, Mark Hudnall, Marvin Zhang, Marwan Aljubeh, Mateusz Litwin, Matthew Zeng, Max Johnson, Maya Shetty, Mayank Gupta, Meghan Shah, Mehmet Yatbaz, Meng~Jia Yang, Mengchao Zhong, Mia Glaese, Mianna Chen, Michael Janner, Michael Lampe, Michael Petrov, Michael Wu, Michele Wang, Michelle Fradin, Michelle Pokrass, Miguel Castro, Miguel Oom~Temudo de~Castro, Mikhail Pavlov, Miles Brundage, Miles Wang, Minal
  Khan, Mira Murati, Mo~Bavarian, Molly Lin, Murat Yesildal, Nacho Soto, Natalia Gimelshein, Natalie Cone, Natalie Staudacher, Natalie Summers, Natan LaFontaine, Neil Chowdhury, Nick Ryder, Nick Stathas, Nick Turley, Nik Tezak, Niko Felix, Nithanth Kudige, Nitish Keskar, Noah Deutsch, Noel Bundick, Nora Puckett, Ofir Nachum, Ola Okelola, Oleg Boiko, Oleg Murk, Oliver Jaffe, Olivia Watkins, Olivier Godement, Owen Campbell-Moore, Patrick Chao, Paul McMillan, Pavel Belov, Peng Su, Peter Bak, Peter Bakkum, Peter Deng, Peter Dolan, Peter Hoeschele, Peter Welinder, Phil Tillet, Philip Pronin, Philippe Tillet, Prafulla Dhariwal, Qiming Yuan, Rachel Dias, Rachel Lim, Rahul Arora, Rajan Troll, Randall Lin, Rapha~Gontijo Lopes, Raul Puri, Reah Miyara, Reimar Leike, Renaud Gaubert, Reza Zamani, Ricky Wang, Rob Donnelly, Rob Honsby, Rocky Smith, Rohan Sahai, Rohit Ramchandani, Romain Huet, Rory Carmichael, Rowan Zellers, Roy Chen, Ruby Chen, Ruslan Nigmatullin, Ryan Cheu, Saachi Jain, Sam Altman, Sam Schoenholz, Sam
  Toizer, Samuel Miserendino, Sandhini Agarwal, Sara Culver, Scott Ethersmith, Scott Gray, Sean Grove, Sean Metzger, Shamez Hermani, Shantanu Jain, Shengjia Zhao, Sherwin Wu, Shino Jomoto, Shirong Wu, Shuaiqi, Xia, Sonia Phene, Spencer Papay, Srinivas Narayanan, Steve Coffey, Steve Lee, Stewart Hall, Suchir Balaji, Tal Broda, Tal Stramer, Tao Xu, Tarun Gogineni, Taya Christianson, Ted Sanders, Tejal Patwardhan, Thomas Cunninghman, Thomas Degry, Thomas Dimson, Thomas Raoux, Thomas Shadwell, Tianhao Zheng, Todd Underwood, Todor Markov, Toki Sherbakov, Tom Rubin, Tom Stasi, Tomer Kaftan, Tristan Heywood, Troy Peterson, Tyce Walters, Tyna Eloundou, Valerie Qi, Veit Moeller, Vinnie Monaco, Vishal Kuo, Vlad Fomenko, Wayne Chang, Weiyi Zheng, Wenda Zhou, Wesam Manassra, Will Sheu, Wojciech Zaremba, Yash Patil, Yilei Qian, Yongjik Kim, Youlong Cheng, Yu~Zhang, Yuchen He, Yuchen Zhang, Yujia Jin, Yunxing Dai, and Yury Malkov. 2024.
\newblock \href {http://arxiv.org/abs/2410.21276} {Gpt-4o system card}.

\bibitem[{Partanen et~al.(2019)Partanen, H{\"a}m{\"a}l{\"a}inen, and Alnajjar}]{finnish}
Niko Partanen, Mika H{\"a}m{\"a}l{\"a}inen, and Khalid Alnajjar. 2019.
\newblock \href {https://doi.org/10.18653/v1/D19-5519} {Dialect text normalization to normative standard {F}innish}.
\newblock In \emph{Proceedings of the 5th Workshop on Noisy User-generated Text (W-NUT 2019)}, pages 141--146, Hong Kong, China. Association for Computational Linguistics.

\bibitem[{Pavlopoulos et~al.(2024)Pavlopoulos, Louridas, and Filos}]{pavlopoulos-etal-2024-towards}
John Pavlopoulos, Panos Louridas, and Panagiotis Filos. 2024.
\newblock \href {https://doi.org/10.18653/v1/2024.emnlp-main.661} {Towards a {G}reek proverb atlas: Computational spatial exploration and attribution of {G}reek proverbs}.
\newblock In \emph{Proceedings of the 2024 Conference on Empirical Methods in Natural Language Processing}, pages 11842--11854, Miami, Florida, USA. Association for Computational Linguistics.

\bibitem[{Pavlopoulos et~al.(2021)Pavlopoulos, Sorensen, Laugier, and Androutsopoulos}]{pavlopoulos-etal-2021-semeval}
John Pavlopoulos, Jeffrey Sorensen, L{\'e}o Laugier, and Ion Androutsopoulos. 2021.
\newblock \href {https://doi.org/10.18653/v1/2021.semeval-1.6} {{S}em{E}val-2021 task 5: Toxic spans detection}.
\newblock In \emph{Proceedings of the 15th International Workshop on Semantic Evaluation (SemEval-2021)}, pages 59--69, Online. Association for Computational Linguistics.

\bibitem[{Ramponi and Casula(2023)}]{ramponi-casula-2023-geolingit}
Alan Ramponi and Camilla Casula. 2023.
\newblock {G}eo{L}ing{I}t at {EVALITA} 2023: Overview of the geolocation of linguistic variation in {I}taly task.
\newblock In \emph{Proceedings of the Eighth Evaluation Campaign of Natural Language Processing and Speech Tools for Italian. Final Workshop (EVALITA 2023)}, Parma, Italy. CEUR.org.

\bibitem[{Scherrer and Ljubeši{\'c}(2016)}]{swiss}
Yves Scherrer and Nikola Ljubeši{\'c}. 2016.
\newblock \href {https://api.semanticscholar.org/CorpusID:67326182} {Automatic normalisation of the {S}wiss {G}erman {A}rchi{M}ob corpus using character-level machine translation}.
\newblock In \emph{Conference on Natural Language Processing}.

\bibitem[{Trudgill(2003)}]{dialects}
Peter Trudgill. 2003.
\newblock \href {https://doi.org/https://doi.org/10.1075/jgl.4.04tru} {Modern greek dialects: A preliminary classification}.
\newblock \emph{Journal of Greek Linguistics}, 4(1):45--63.

\end{thebibliography}

\appendix

\section{Dialect Groups}
\label{sec:groups}

\subsection{Northern}
This includes: Macedonia, Thrace, Eastern Thrace, Skyros, Epirus, Ioannina, Asia Minor, Aetolia, Euboea and Lesbos.

\subsection{Southern}
This includes: Amorgos, Arcadia, Achaea, Ionian Islands, Thesprotia, Karpathos, Cephalonia, Crete, Cyprus, Laconia, Naxos and Rhodes.

\subsection{Pontus}
This includes Pontus, a very divergent dialect which doesn't share many features with the others.

\section{Major Changes per Dialect Group}
\label{sec:rules}

Below is one major example of the changes our scripts make for each group:

\subsection{Northern}

A major characteristic of Northern dialects is ``Northern vocalism'', which raises standard mid vowels (/o/, /e/) to high vowels (/u/, /i/) in unstressed positions, while original high vowels disappear under the same circumstances. Completely undoing this rule is difficult, as it is facultative and therefore not reversible. However, there are certain patterns, such as the word /u/ followed by another ending in unstressed /-us/, which are almost guaranteed to be the result of this rule, and are therefore safe to reverse to /o/ and /-os/ at this stage.

\subsection{Southern}

A feature of Southern dialects is the palatalization of velars, especially /k/, before vowels (/e/, /i/). The resulting palatal is represented differently in each dialect due to the decisions of each transcriber who happened to collect data from each region. Similarly to above, it is difficult to know which palatal was original or resulted from this rule, so the process is not completely reversible, but we revert it in specific cases where it is almost certain.

\subsection{Pontus}

Pontic Greek uses /do/ in place of Standard Modern Greek /ti/ (meaning ``what''), while in other dialects this usually represents a voiced version of the definite article.

\section{Full Prompt Templates}
\label{sec:templates}

We used the following three prompt templates, one for each group of Greek dialects. ``<place>'' is replaced by the area label, while ``<text>'' is replaced by the source dialectal proverb.

\subsection{Northern}

\begin{quote} 
\centering 
`Given a Greek sentence from <place>. Translate it to standard Greek. Keep the same style, do not make it more official. Use words with the same etymology if and only if they exist in standard Greek, otherwise use different words. Show just the translation and nothing else.

            For example:

            <place>:
            \foreignlanguage{greek}{Γίδα ψουριάρα, νουρά κουρδουμέν'}

            Standard Greek:
            \foreignlanguage{greek}{Γίδα ψωριάρα, ουρά κορδωμένη}

            <place>:
            \foreignlanguage{greek}{Μι πήρι, σι πήρι, τουν πήρι του πουτάμ'}

            Standard Greek:
            \foreignlanguage{greek}{Με πήρε, σε πήρε, τον πήρε το ποτάμι}

            <place>:
            \foreignlanguage{greek}{Τ' γάμσι του κέρατου}

            Standard Greek:
            \foreignlanguage{greek}{Του γάμησε το κέρατο}

            <place>:
            <text>

            Standard Greek:' 
\end{quote}

\subsection{Southern}

\begin{quote} 
\centering 
`Given a Greek sentence from <place>. Translate it to standard Greek. Keep the same style, do not make it more official. Use words with the same etymology if and only if they exist in standard Greek, otherwise use different words. Show just the translation and nothing else.

For example:

<place>:
\foreignlanguage{greek}{Καλλιά 'ν' το διακονίκι, παρά το βασιλίκι}

Standard Greek:
\foreignlanguage{greek}{Καλύτερα είναι το διακονίκι, παρά το βασιλίκι}

<place>:
\foreignlanguage{greek}{Τάχει η γραι στο λοϊσμό τζη τα θωρεί και στο όνειρό τζη}

Standard Greek:
\foreignlanguage{greek}{Τά 'χει η γρια στον λογισμό της τα βλέπει και στο όνειρό της}

<place>:
\foreignlanguage{greek}{Των βρενίμων τα παιδκιά πριν πεινασουν μαειρεύκουν}

Standard Greek:
\foreignlanguage{greek}{Των φρονίμων τα παιδιά πριν πεινάσουν μαγειρεύουν}

<place>:
<proverb>

Standard Greek:' 
\end{quote}

\subsection{Pontus}

\begin{quote} 
\centering 
`Given a Greek sentence from \foreignlanguage{greek}{Πόντος}. Translate it to standard Greek. Keep the same style, do not make it more official. Use words with the same etymology if and only if they exist in standard Greek, otherwise use different words. Show just the translation and nothing else.

For example:

\foreignlanguage{greek}{Πόντος}:
\foreignlanguage{greek}{Ποιος βάλλ' το χέρ΄ν ατ' 'ς σο μέλ' και 'κι λείχ' τα δάχτυλα 'τ'}

Standard Greek:
\foreignlanguage{greek}{Ποιος βάζει το χέρι του στο μέλι και δεν γλείφει τα δάχτυλά του}

\foreignlanguage{greek}{Πόντος}:
\foreignlanguage{greek}{Κι'αν παθάνης κι μαθάνεις}

Standard Greek:
\foreignlanguage{greek}{Αν δεν παθαίνεις δεν μαθαίνεις}

\foreignlanguage{greek}{Πόντος}:
\foreignlanguage{greek}{Ο νέον θολόν ποτάμιν είναι!}

Standard Greek:
\foreignlanguage{greek}{Ο νέος θολό ποτάμι είναι!}

\foreignlanguage{greek}{Πόντος}:
<proverb>

Standard Greek:' 
\end{quote}

\section{Detailed Annotation Statistics}
\label{sec:stats}

\subsection{Pearson Correlations}

We report the average pairwise Pearson Correlation for the ratings of the outputs of each model among the three annotators.

Numbers closer to 1 indicate better correlation.

\needspace{10\baselineskip}
\subsubsection{Form}
\FloatBarrier
\begin{table}[h]
\centering
\setlength{\tabcolsep}{0pt}
\begin{tabular}{clc}
\toprule
& \textbf{Model} & \textbf{Pearson}\\
\midrule
&\verb|GPT 3s+RBN| & {0.733} \\
&\verb|GPT 3s| & {0.822} \\
&\verb|Llama 3s+RBN| & {0.601} \\ 
&\verb|Llama 9s| & {0.787} \\ 
\bottomrule
\end{tabular}
\caption{Average Pearson Correlation for each model.}
\end{table}
\FloatBarrier

\needspace{15\baselineskip}
\subsubsection{Meaning}
\begin{table}[h]
\centering
\setlength{\tabcolsep}{0pt}
\begin{tabular}{clc}
\toprule
& \textbf{Model} & \textbf{Pearson}\\
\midrule
&\verb|GPT 3s+RBN| & {0.646} \\
&\verb|GPT 3s| & {0.731} \\
&\verb|Llama 3s+RBN| & {0.821} \\ 
&\verb|Llama 9s| & {0.762} \\ 
\bottomrule
\end{tabular}
\caption{Average Pearson Correlation for each model.}
\end{table}
\FloatBarrier

\subsection{Intraclass Correlation Coefficients}

We specifically report the ICC (2,k) statistic, calculated for the average of ratings provided by a set of annotators, where the annotators are treated as random effects under a two-way random effects model. This is because we use the average of their evaluations in our analyses instead of any specific individual rating, while our annotators are used as representatives of the Greek-speaking population, and we are interested in their evaluations as part of this group.

Numbers closer to 1 indicate better correlation, with 0.75 to 0.9 generally considered good, and higher than 0.90 excellent \cite{icc}.
\subsubsection{Form}

\begin{table}[h]
\centering
\small
\setlength{\tabcolsep}{1pt}
\begin{tabular}{llcccccl}
\toprule
\textbf{Model}            & \textbf{ICC} & \textbf{F}  & \textbf{df1} & \textbf{df2} & \textbf{p} & \textbf{CI95\%} \\ 
\midrule
\verb|GPT 3s+RBN| & 0.884  & 8.819 & 26  & 52  & $2.24\times10^{-11}$       & [0.78, 0.94] \\
\verb|GPT 3s| & 0.934  & 14.700 & 26  & 52  & $6.77\times10^{-16}$       & [0.87, 0.97] \\
\verb|Llama 3s+RBN| & 0.790  & 5.201 & 26  & 52  & $2.24\times10^{-7}$       & [0.60, 0.90] \\
\verb|Llama 9s| & 0.888  & 11.264 & 26  & 52  & $1.79\times10^{-13}$       & [0.76, 0.95] \\
\bottomrule
\end{tabular}
\caption{ICC (2,k)  and the associated F-statistic, numerator (df1) and denominator (df2) degrees of freedom, p-value (for the possibility of the true ICC being 0) and 95\% confidence interval for the form ratings of each model.}
\end{table}
\FloatBarrier
\needspace{10\baselineskip}
\subsubsection{Meaning}

\begin{table}[h]
\centering
\small
\setlength{\tabcolsep}{1pt}
\begin{tabular}{llcccccl}
\toprule
\textbf{Model}            & \textbf{ICC} & \textbf{F}  & \textbf{df1} & \textbf{df2} & \textbf{p} & \textbf{CI95\%} \\ 
\midrule
\texttt{GPT 3s+RBN} & 0.783  & 4.667 & 26  & 52  & 1.00$\times10^{-6}$       & [0.59, 0.89] \\
\texttt{GPT 3s} & 0.893  & 9.065 & 26  & 52  & 1.32$\times10^{-11}$       & [0.80, 0.95] \\
\texttt{Llama 3s+RBN} & 0.910  & 12.133 & 26  & 52  & 3.90$\times10^{-14}$       & [0.83, 0.96] \\
\texttt{Llama 9s} & 0.875  & 9.679 & 26  & 52  & 3.71$\times10^{-12}$     & [0.75, 0.94] \\
\bottomrule
\end{tabular}
\caption{ICC (2,k)  and the associated F-statistic, numerator (df1) and denominator (df2) degrees of freedom, p-value (for the possibility of the true ICC being 0) and 95\% confidence interval for the meaning ratings of each model.}
\end{table}

\FloatBarrier
\subsection{Paired T-Tests for Statistical Significance}

We report on the statistical significance of each model's score being higher than the following in the sequence in which they were ranked.

P-values \(<0.05\) are typically considered statistically significant.

\needspace{10\baselineskip}
\subsubsection{Form}

\begin{table}[h]
\centering
\setlength{\tabcolsep}{0pt}
\begin{tabular}{clcc}
\toprule
& \textbf{Model} & \textbf{t-statistic} & \textbf{p-value}\\
\midrule
&\verb|GPT (3s+RBN - 3s)| & {2.083} & {0.041} \\
&\verb|GPT 3s - Llama 3s+RBN| & {9.385} & {\(1.9\times10^{-14}\)} \\
&\verb|Llama (3s+RBN - 9s)| & {3.295} & {0.001} \\ 
\bottomrule
\end{tabular}
\caption{Statistical significance of each model's form-score being higher than the following in the sequence in which they were ranked. All p-values are \(<0.05\).}
\end{table}

\FloatBarrier
\subsubsection{Meaning}

\begin{table}[h]
\centering
\setlength{\tabcolsep}{0pt}
\begin{tabular}{clcc}
\toprule
& \textbf{Model} & \textbf{t-statistic} & \textbf{p-value}\\
\midrule
&\verb|GPT (3s+RBN - 3s)| & {3.202} & {0.002} \\
&\verb|GPT 3s - Llama 3s+RBN| & {7.157} & {\(3.9\times10^{-10}\)} \\
&\verb|Llama (3s+RBN - 9s)| & {3.373} & {0.001} \\ 
\bottomrule
\end{tabular}
\caption{Statistical significance of each model's meaning-score being higher than the following in the sequence in which they were ranked. All p-values are \(<0.05\).}
\end{table}

\FloatBarrier
\section{Demographic details of the annotators}
Out of our three annotators, two were native speakers of Standard Modern Greek and the Cretan dialect, while the other was a native speaker of Standard Modern Greek and Northern Greek. The contents of our dataset are generally understandable to all Greek speakers.

\FloatBarrier
\needspace{15\baselineskip}
\section{Results of K-means for 2 clusters (full size)}
\label{app:kmeans}

\begin{figure}[h]
    \centering
    \includegraphics[width=\linewidth]{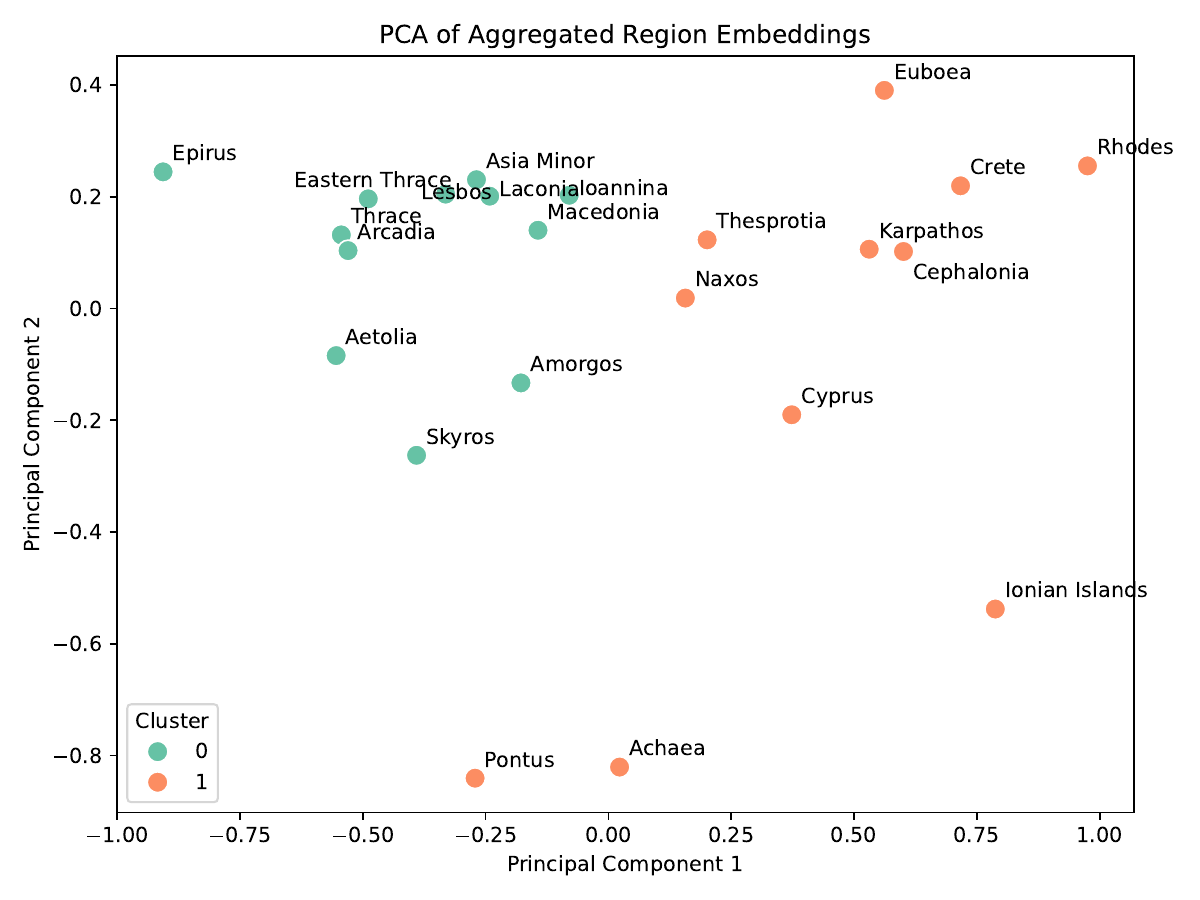} 
    \caption{K-means clustering for 2 clusters using normalized data}
\end{figure}

\begin{figure}[h]
    \centering
    \includegraphics[width=\linewidth]{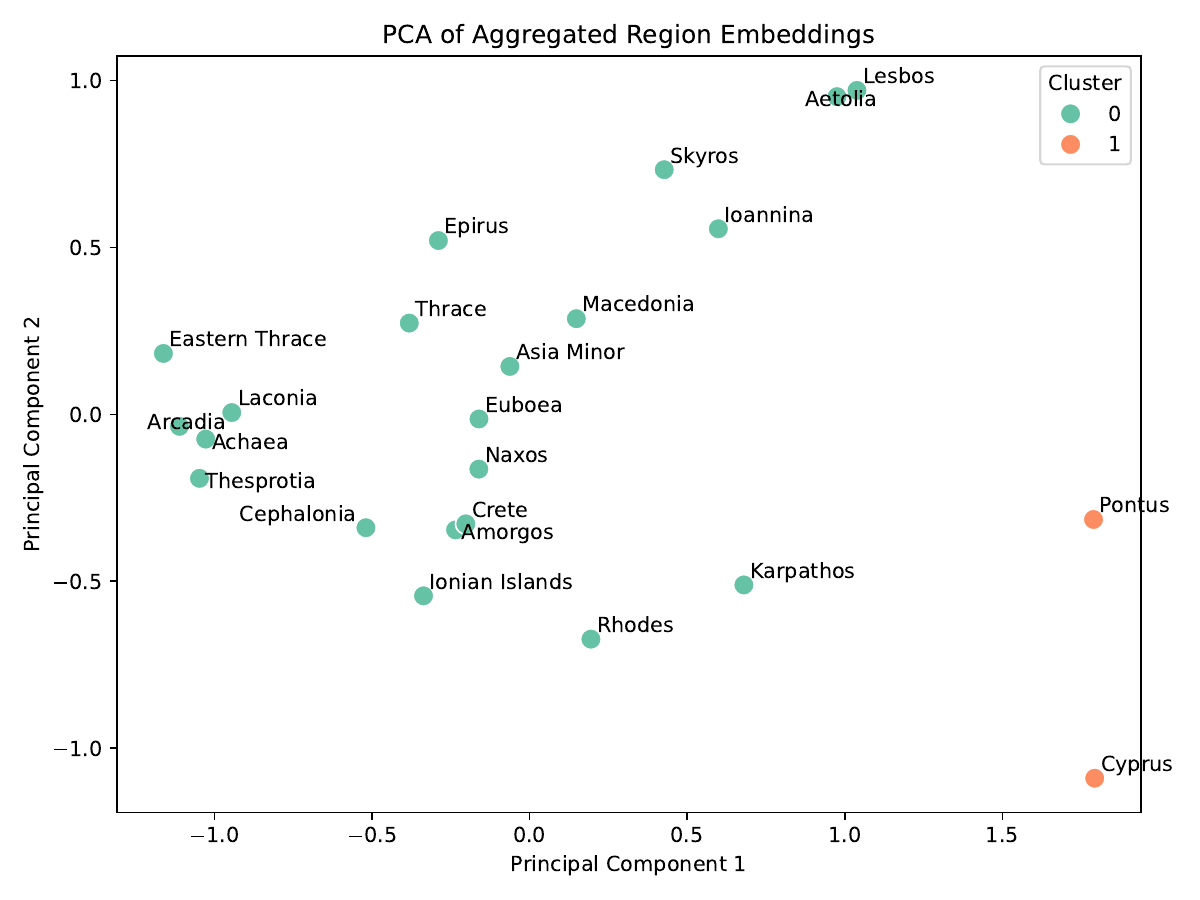} 
    \caption{K-means clustering for 2 clusters using original dialectal data}
\end{figure}

\FloatBarrier
\needspace{120\baselineskip}
\section{Results of other Clustering Algorithms}
\label{sec:clust}

\begin{figure}[h]
    \centering
    \includegraphics[width=\linewidth]{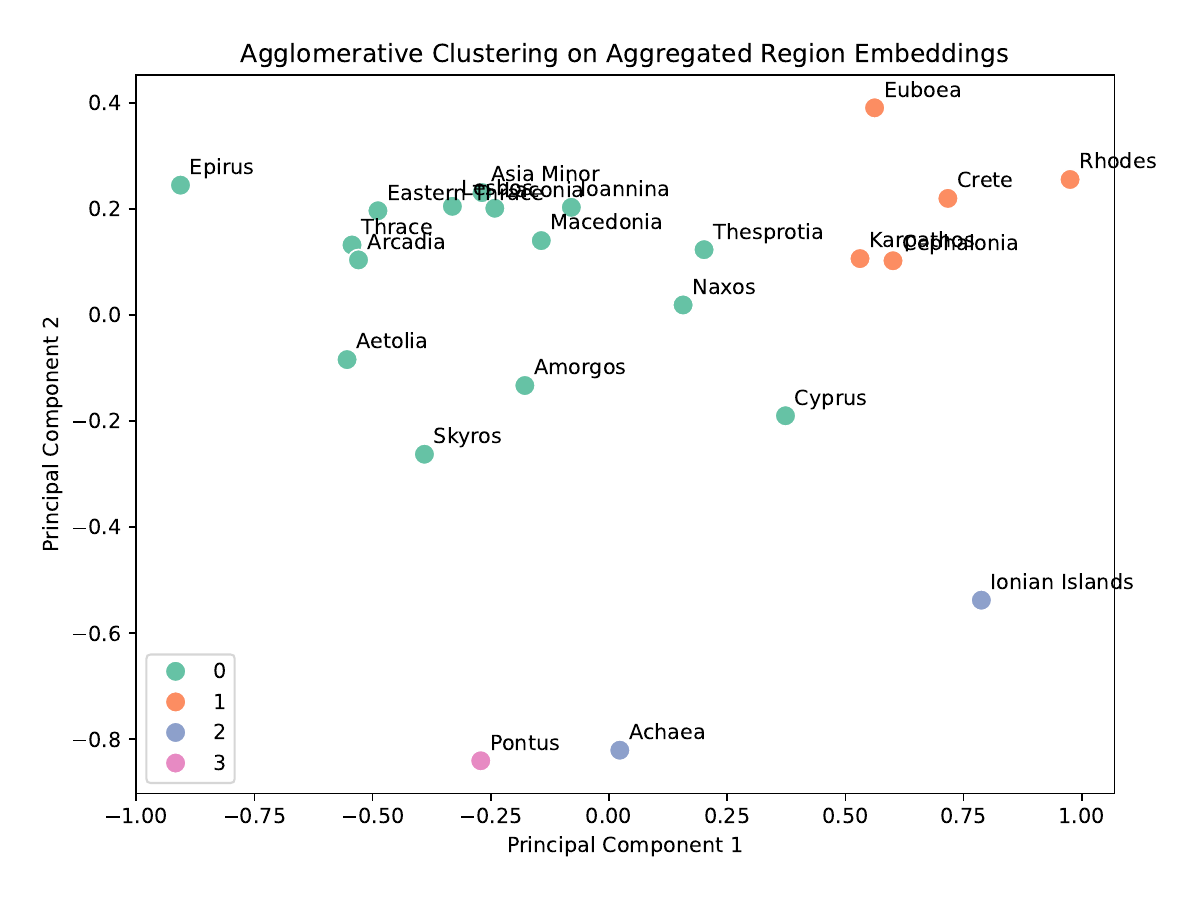} 
    \caption{Agglomerative clustering using normalized data}
\end{figure}

\begin{figure}[h]
    \centering
    \includegraphics[width=\linewidth]{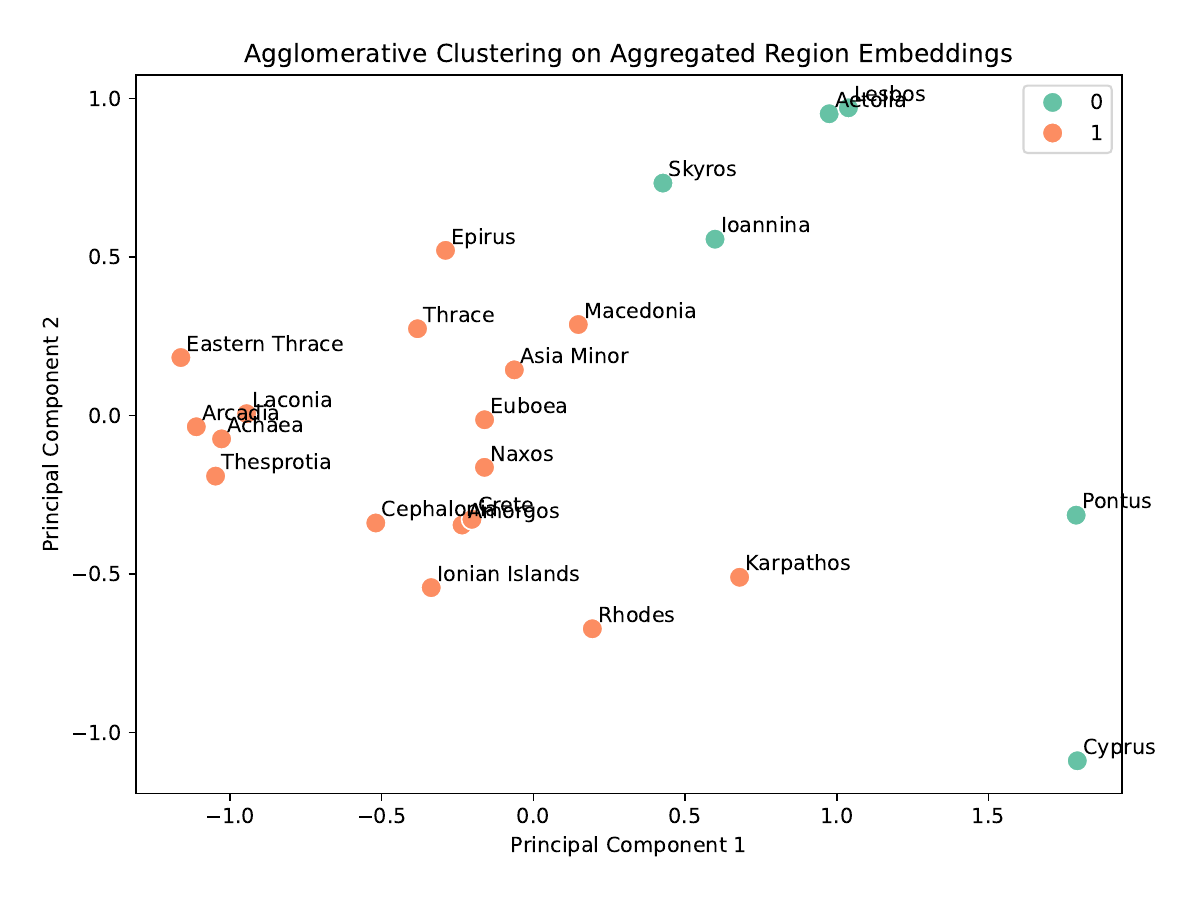} 
    \caption{Agglomerative clustering using original dialectal data}
\end{figure}

\begin{figure}[h]
    \centering
    \includegraphics[width=\linewidth]{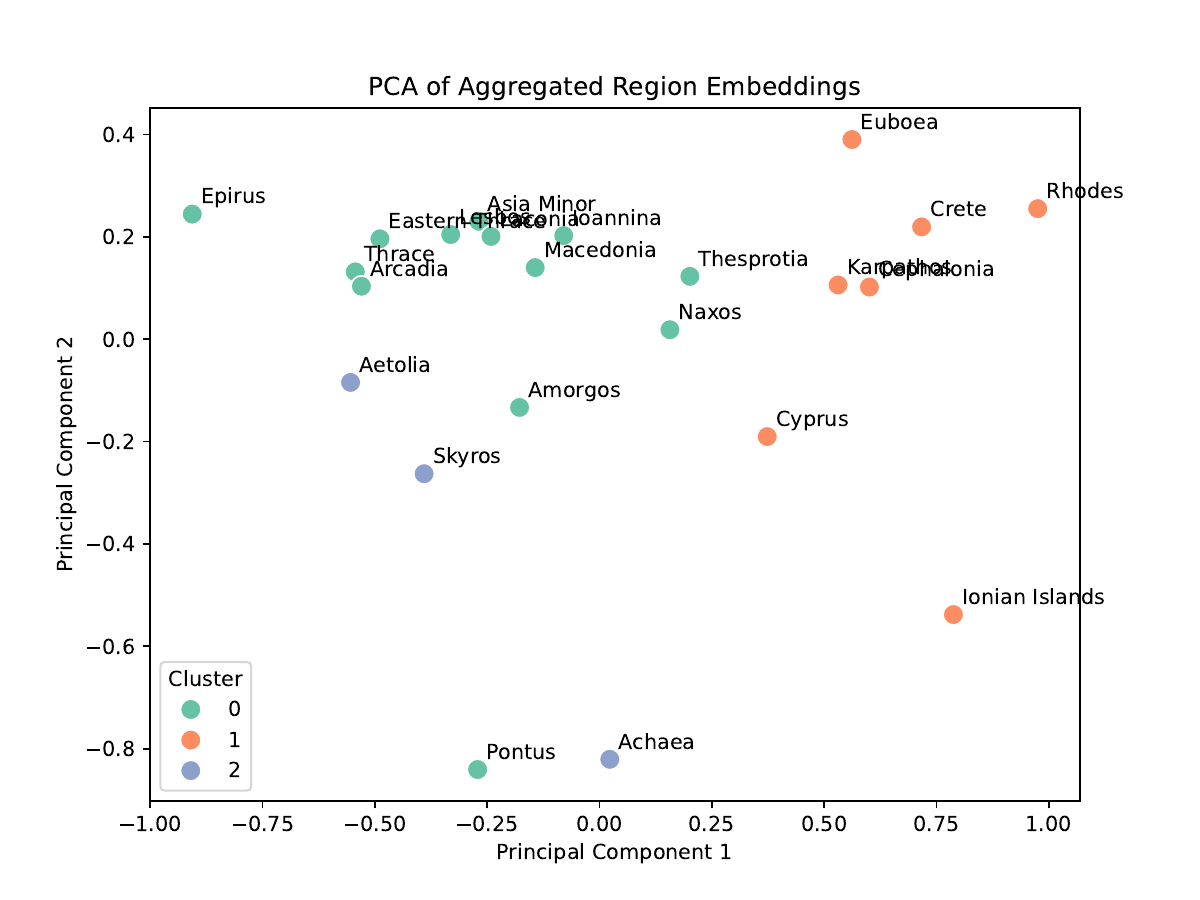} 
    \caption{K-means clustering for 3 clusters using normalized data}
\end{figure}

\begin{figure}[h]
    \centering
    \includegraphics[width=\linewidth]{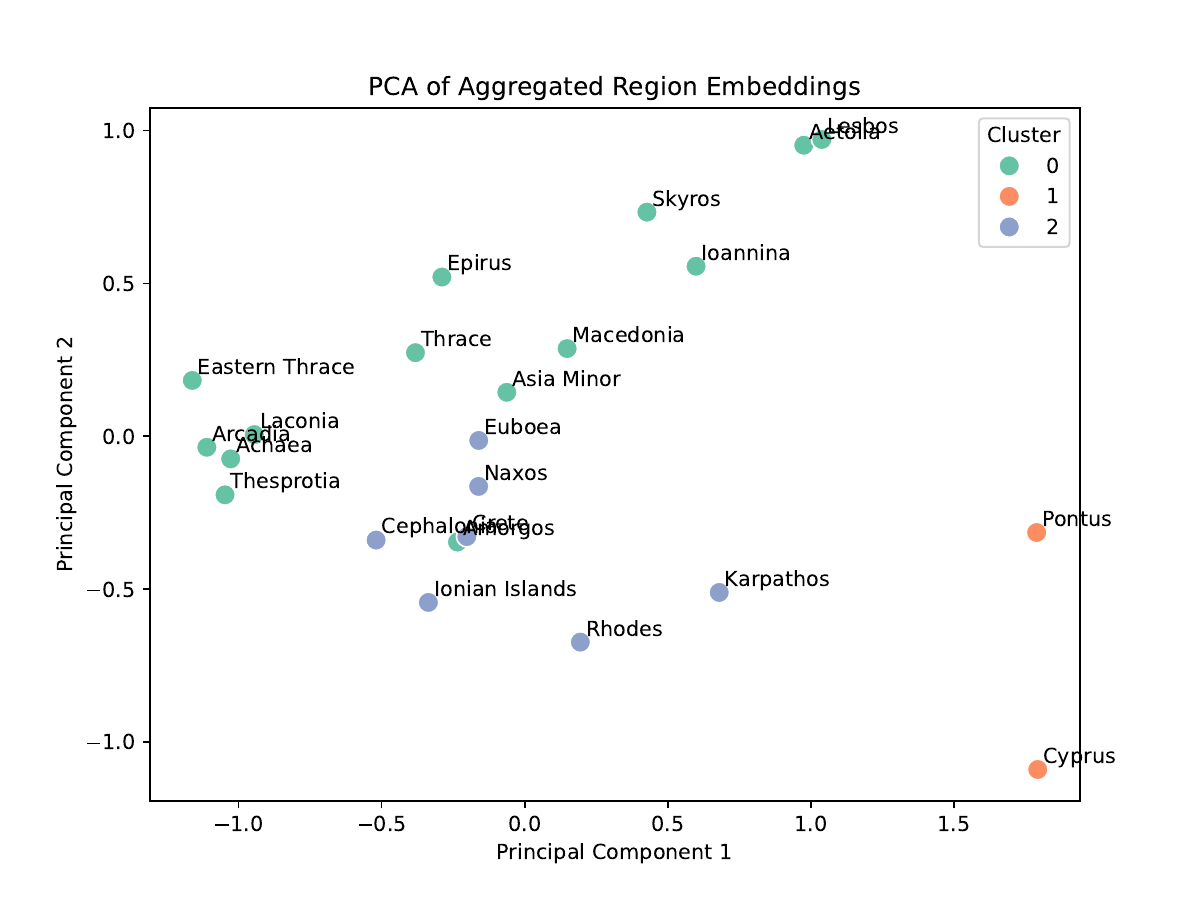} 
    \caption{K-means clustering for 3 clusters using original dialectal data}
\end{figure}

\begin{figure}[h]
    \centering
    \includegraphics[width=\linewidth]{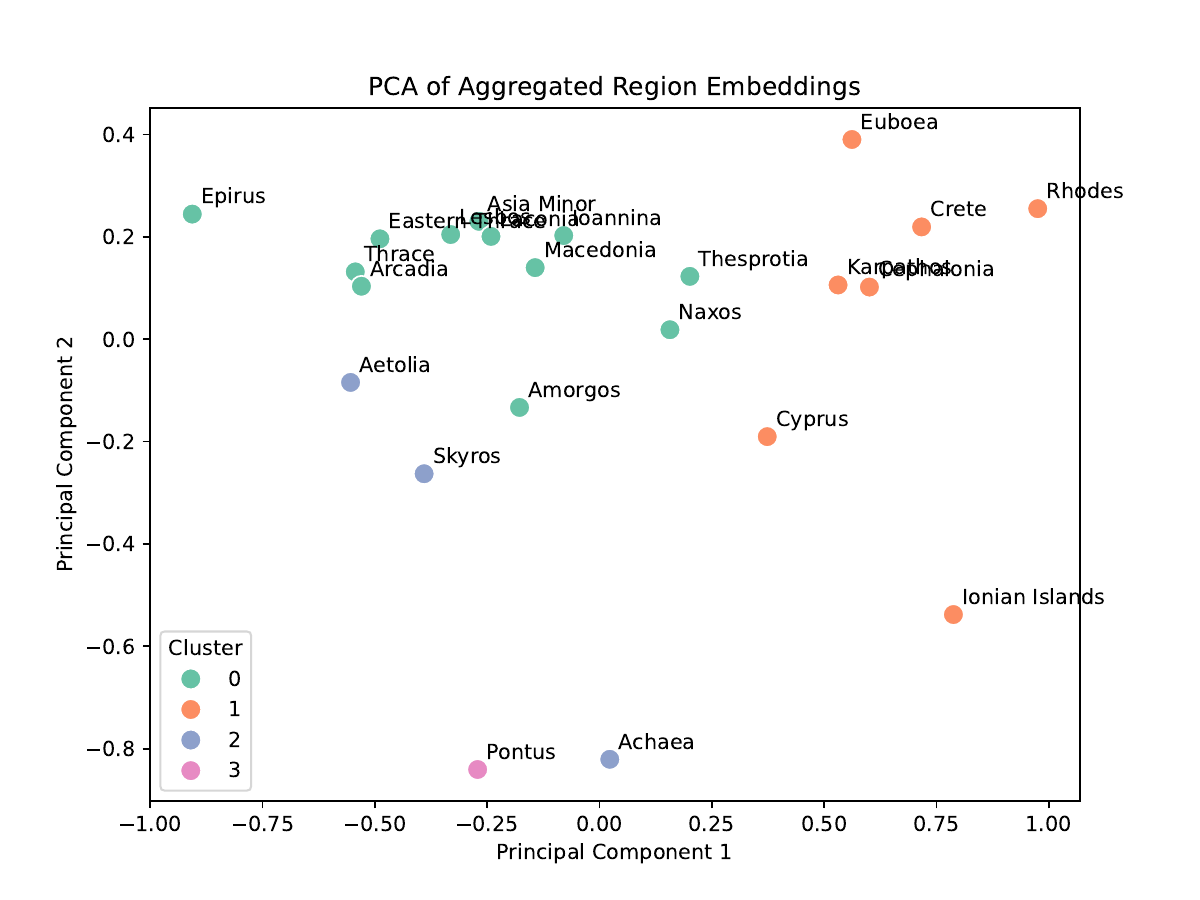} 
    \caption{K-means clustering for 4 clusters using normalized data}
\end{figure}

\begin{figure}[h]
    \centering
    \includegraphics[width=\linewidth]{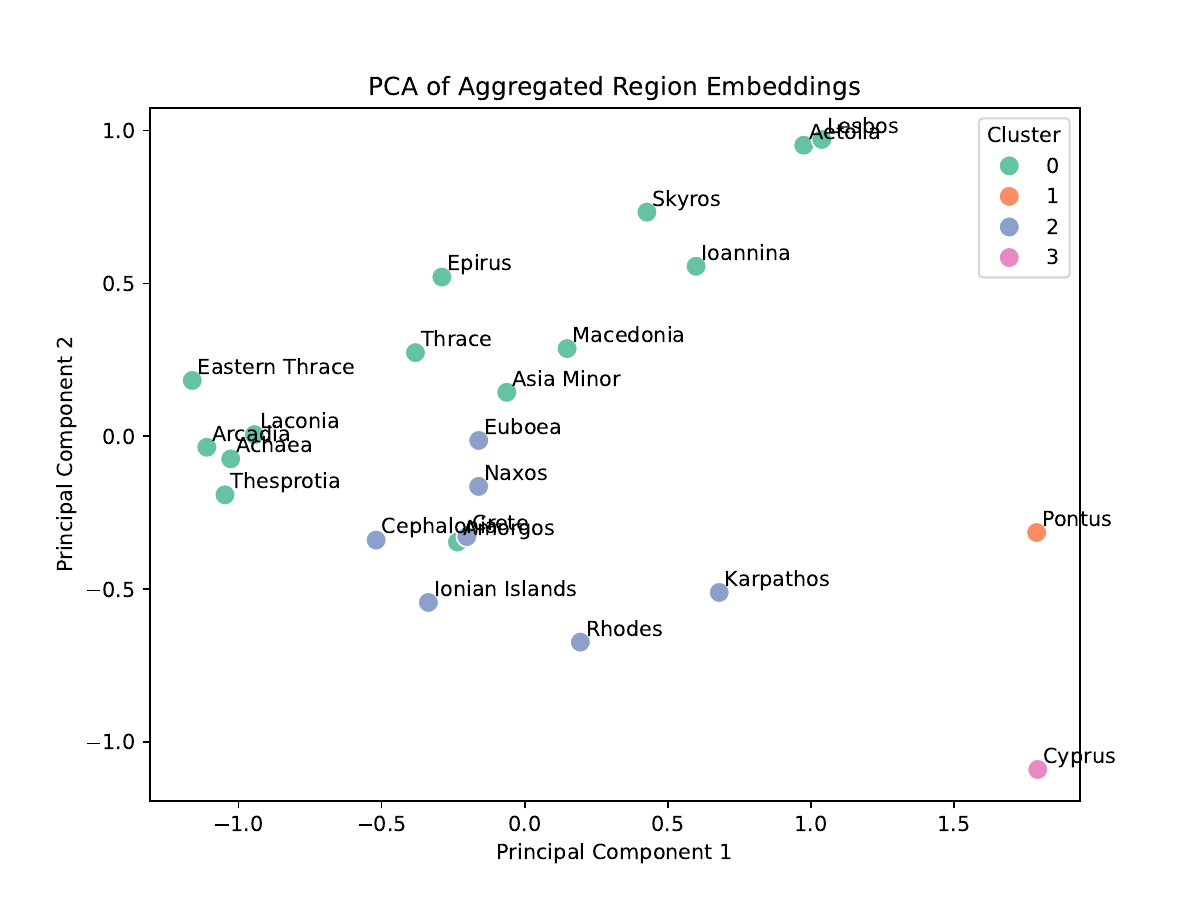} 
    \caption{K-means clustering for 4 clusters using original dialectal data}
\end{figure}

\begin{figure}[h]
    \centering
    \includegraphics[width=\linewidth]{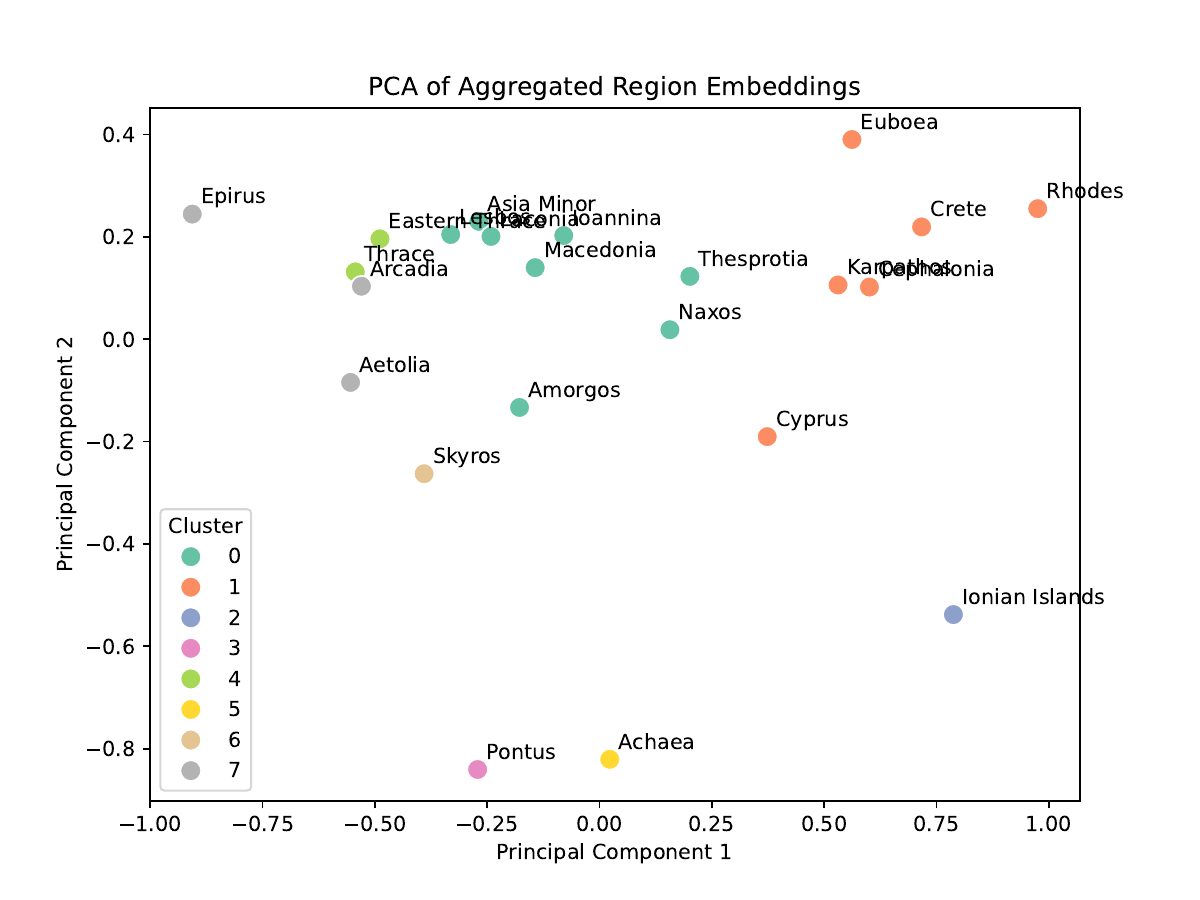} 
    \caption{K-means clustering for 8 clusters using normalized data}
\end{figure}

\begin{figure}[h]
    \centering
    \includegraphics[width=\linewidth]{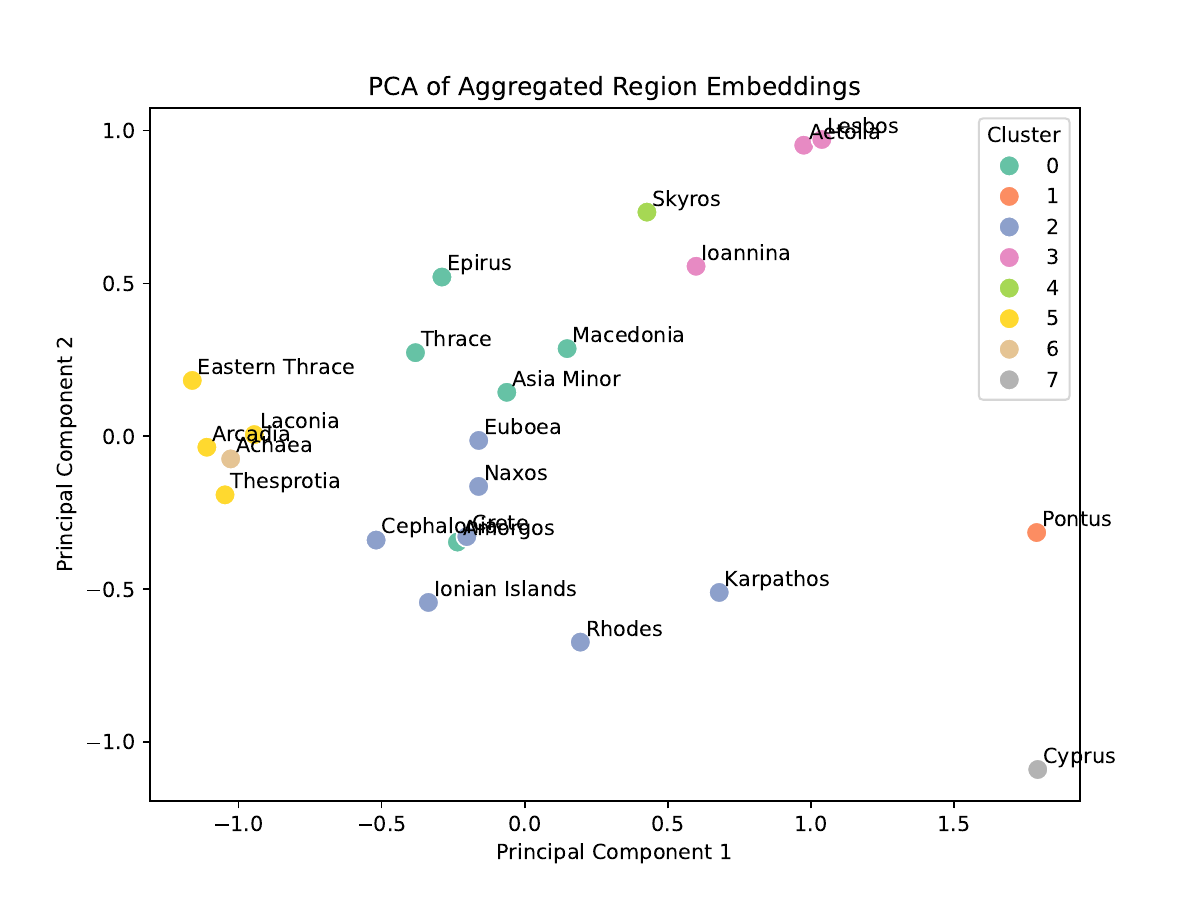} 
    \caption{K-means clustering for 8 clusters using original dialectal data}
\end{figure}

\begin{figure}[h]
    \centering
    \includegraphics[width=\linewidth]{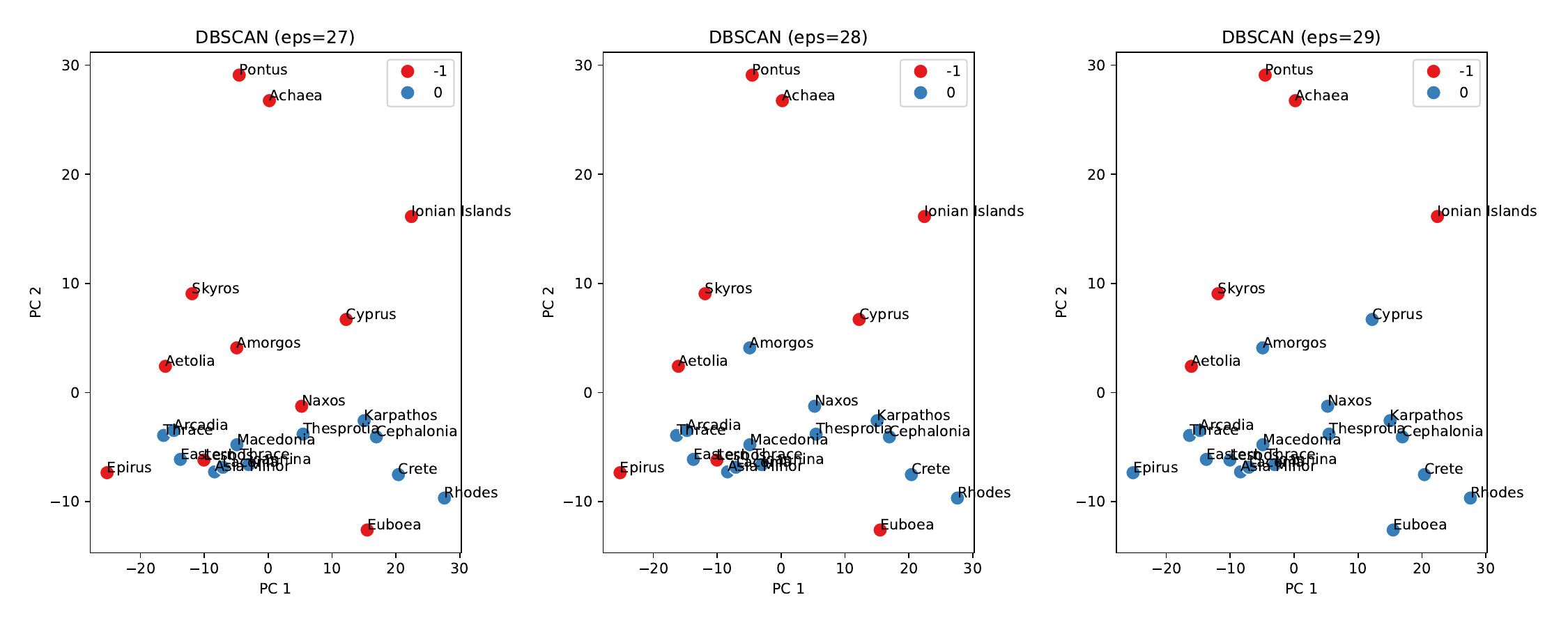} 
    \caption{DBSCAN clustering with various values of eps using normalized data}
\end{figure}

\begin{figure}[h]
    \centering
    \includegraphics[width=\linewidth]{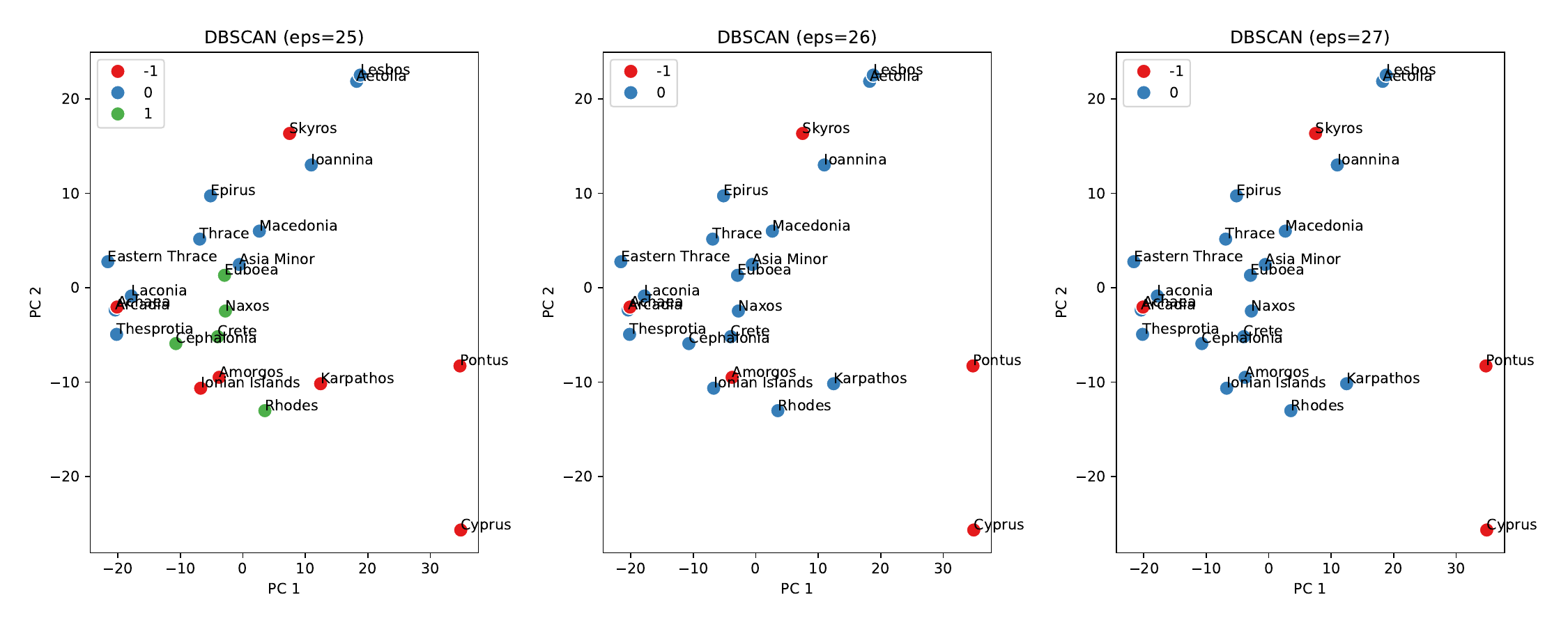} 
    \caption{DBSCAN clustering with various values of eps using original dialectal data}
\end{figure}

\begin{figure}[h]
    \centering
    \includegraphics[width=\linewidth]{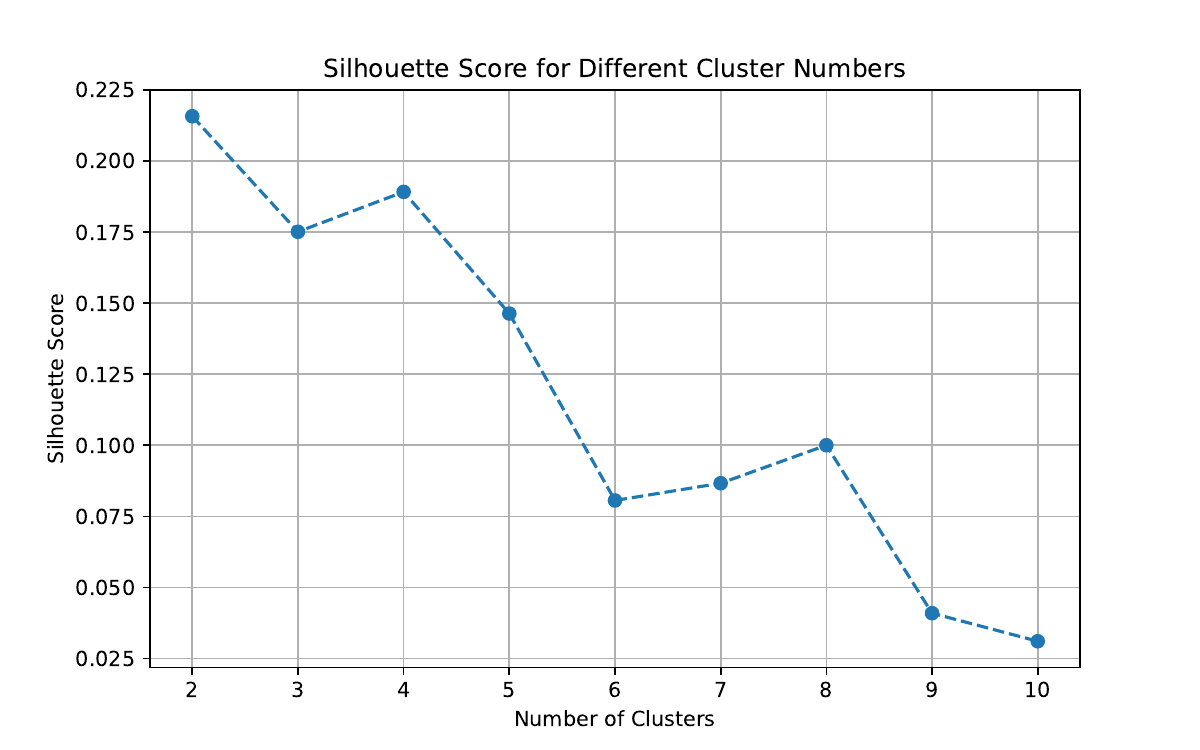} 
    \caption{K-means silhouette score by k to determine optimal number of clusters using normalized data}
\end{figure}

\begin{figure}[h]
    \centering
    \includegraphics[width=\linewidth]{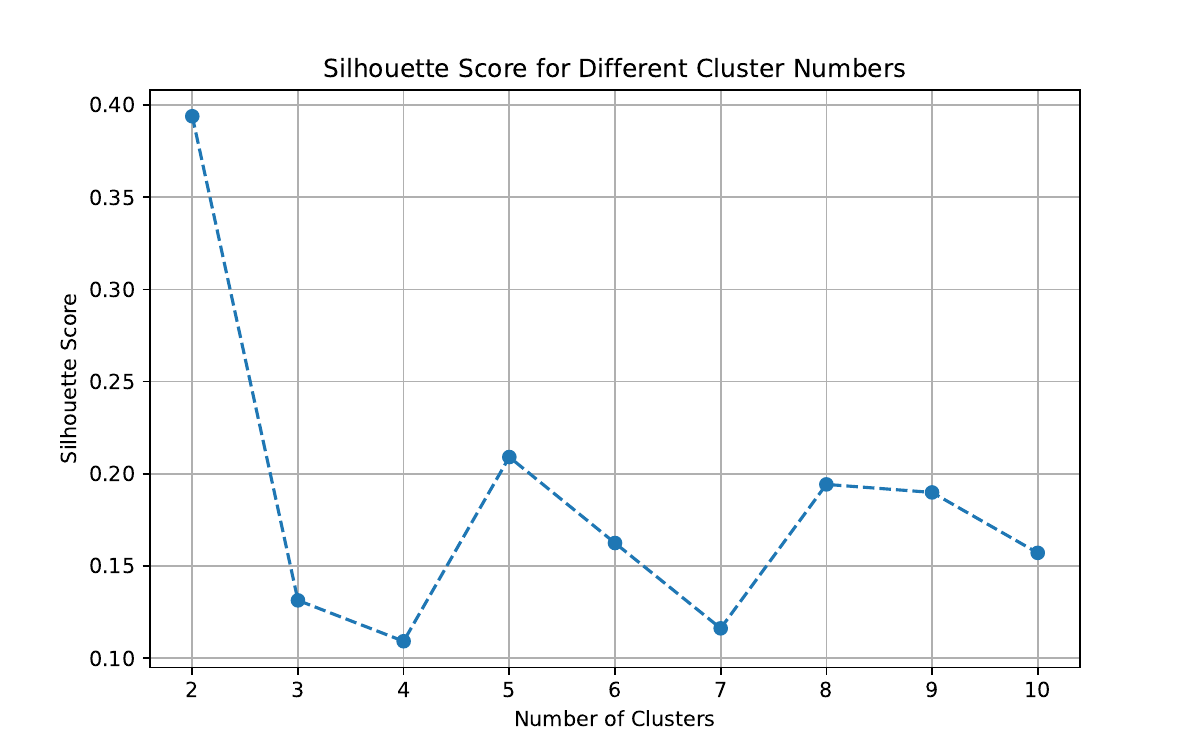} 
    \caption{K-means silhouette score by k to determine optimal number of clusters using original dialectal data}
\end{figure}

\begin{figure}[h]
    \centering
    \includegraphics[width=\linewidth]{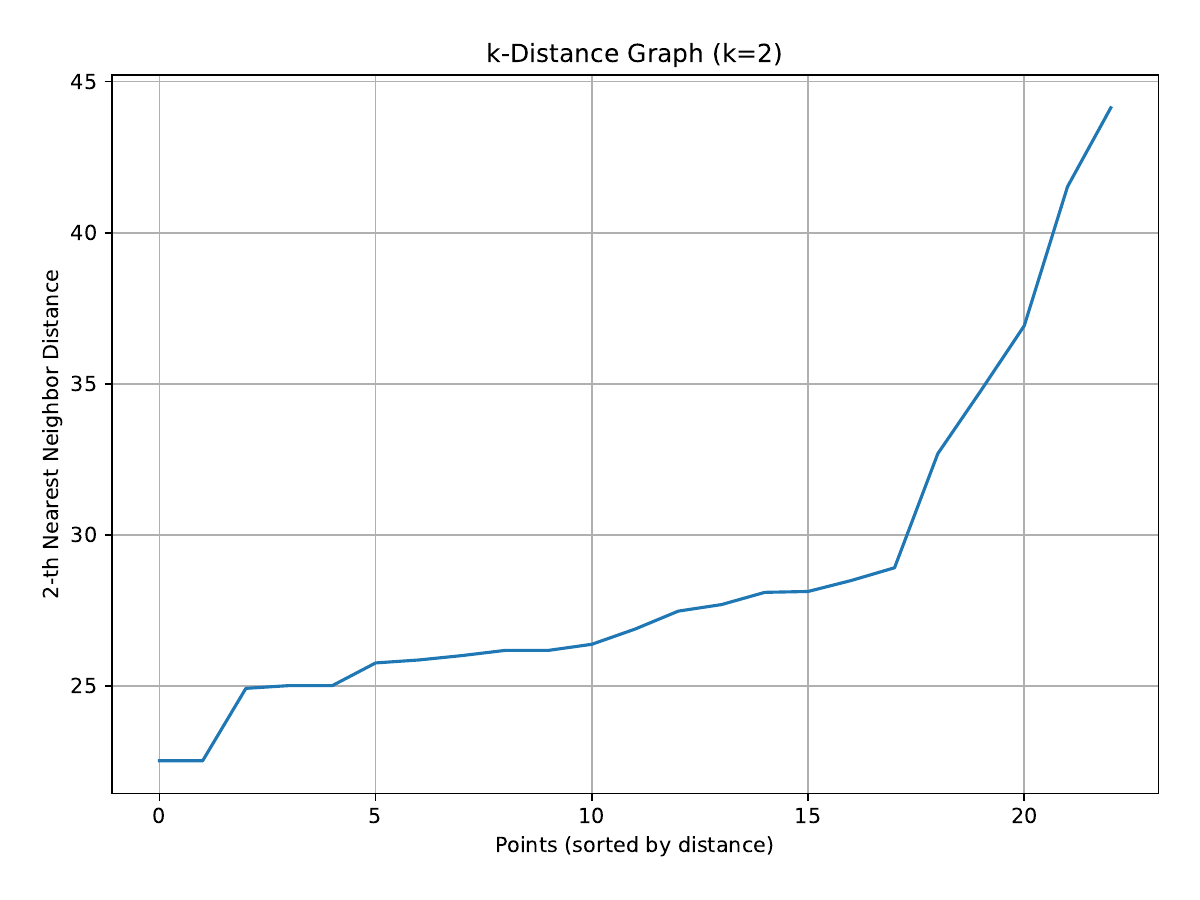} 
    \caption{DBSCAN distance graph for finding the optimal eps parameter through the elbow method using normalized data}
\end{figure}

\begin{figure}[h]
    \centering
    \includegraphics[width=\linewidth]{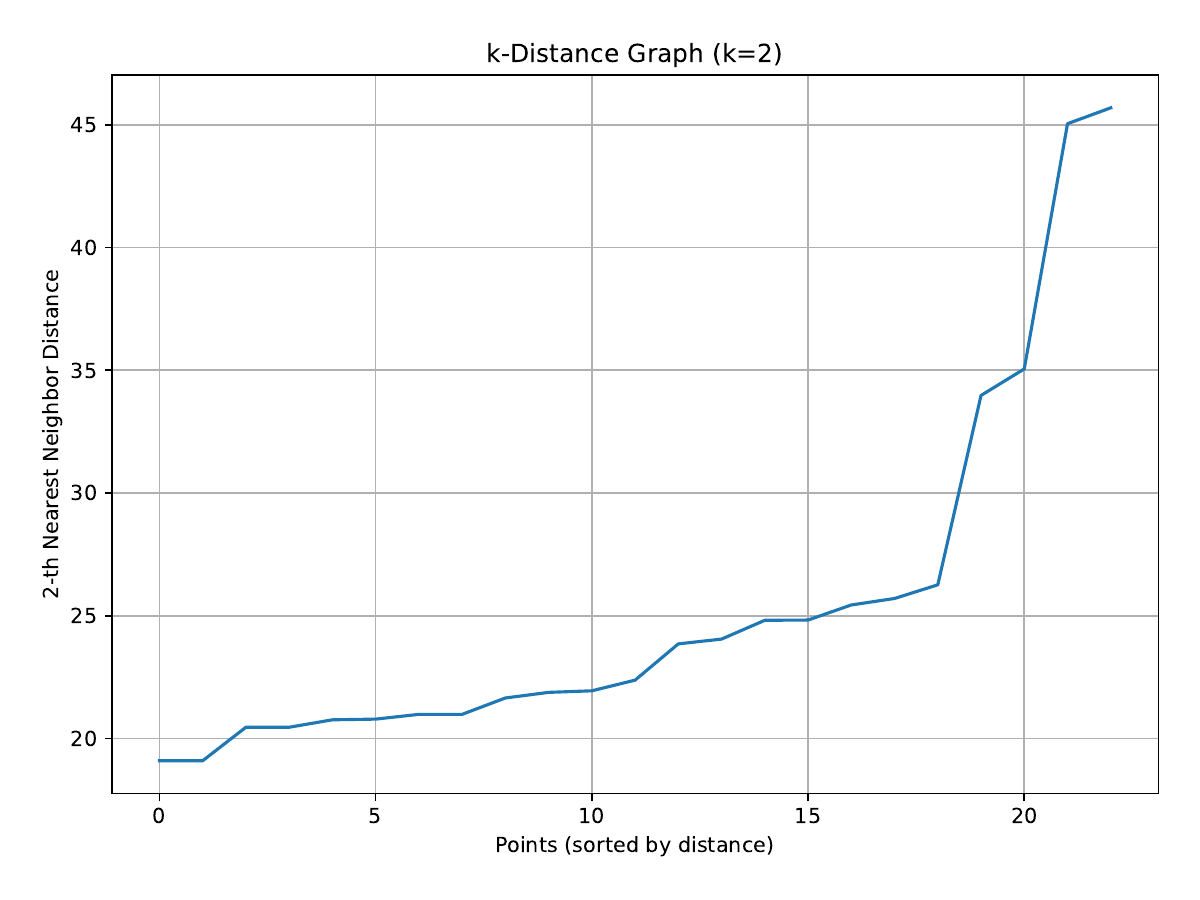} 
    \caption{DBSCAN distance graph for finding the optimal eps parameter through the elbow method using original dialectal data}
\end{figure}

\FloatBarrier
\section{Detailed Results of Downstream Tasks}
\label{sec:results}

\begin{table}[h]
\centering
\setlength{\tabcolsep}{2pt}
\begin{tabular}{clcccc}
\toprule
& \textbf{Model} & \textbf{precision} & \textbf{recall} & \textbf{f1-score} & \textbf{support}\\
\midrule
& \verb|Epirus|       & 0.17 & 0.17 & 0.17 & 23 \\
& \verb|Aetolia|      & 0.38 & 0.58 & 0.46 & 24 \\
& \verb|Amorgos|      & 0.13 & 0.18 & 0.15 & 22 \\
& \verb|Eastern Thrace| & 0.16 & 0.21 & 0.18 & 24 \\
& \verb|Arcadia|      & 0.20 & 0.16 & 0.18 & 31 \\
& \verb|Achaea|       & 0.39 & 0.22 & 0.28 & 32 \\
& \verb|Ionian Islands| & 0.35 & 0.65 & 0.45 & 23 \\
& \verb|Euboea|       & 0.06 & 0.05 & 0.05 & 20 \\
& \verb|Thesprotia|   & 0.05 & 0.05 & 0.05 & 22 \\
& \verb|Thrace|       & 0.25 & 0.16 & 0.20 & 25 \\
& \verb|Ioannina|     & 0.29 & 0.21 & 0.24 & 29 \\
& \verb|Karpathos|    & 0.40 & 0.29 & 0.33 & 28 \\
& \verb|Cephalonia|   & 0.14 & 0.11 & 0.12 & 27 \\
& \verb|Crete|        & 0.35 & 0.27 & 0.30 & 30 \\
& \verb|Cyprus|       & 0.72 & 0.75 & 0.73 & 24 \\
& \verb|Lesbos|       & 0.42 & 0.62 & 0.50 & 24 \\
& \verb|Laconia|      & 0.12 & 0.07 & 0.09 & 27 \\
& \verb|Macedonia|    & 0.37 & 0.26 & 0.30 & 27 \\
& \verb|Asia Minor|   & 0.00 & 0.00 & 0.00 & 18 \\
& \verb|Naxos|        & 0.31 & 0.46 & 0.37 & 24 \\
& \verb|Pontus|       & 0.75 & 0.79 & 0.77 & 19 \\
& \verb|Rhodes|       & 0.26 & 0.23 & 0.24 & 22 \\
& \verb|Skyros|       & 0.45 & 0.60 & 0.51 & 30 \\
\midrule
& \verb|accuracy|     & \multicolumn{1}{c}{} & \multicolumn{1}{c}{} & 0.31 & 575 \\
& \verb|macro avg|    & 0.29 & 0.31 & 0.29 & 575 \\
& \verb|weighted avg| & 0.30 & 0.31 & 0.29 & 575 \\
\bottomrule
\end{tabular}
\caption{Location classification with logistic regression with dialectal data}
\end{table}

\begin{table}[h]
\centering
\setlength{\tabcolsep}{2pt}
\begin{tabular}{clcccc}
\toprule
& \textbf{Model} & \textbf{precision} & \textbf{recall} & \textbf{f1-score} & \textbf{support}\\
\midrule
& \verb|Epirus|       & 0.08 & 0.09 & 0.09 & 23 \\
& \verb|Aetolia|       & 0.16 & 0.12 & 0.14 & 24 \\
& \verb|Amorgos|       & 0.16 & 0.10 & 0.12 & 29 \\
& \verb|Eastern Thrace| & 0.14 & 0.14 & 0.14 & 22 \\
& \verb|Arcadia|       & 0.10 & 0.07 & 0.08 & 28 \\
& \verb|Achaea|         & 0.13 & 0.07 & 0.10 & 27 \\
& \verb|Ionian Islands|     & 0.24 & 0.33 & 0.28 & 30 \\
& \verb|Euboea|       & 0.14 & 0.12 & 0.13 & 24 \\
& \verb|Thesprotia|     & 0.13 & 0.17 & 0.15 & 24 \\
& \verb|Thrace|        & 0.16 & 0.10 & 0.12 & 31 \\
& \verb|Ioannina|     & 0.08 & 0.06 & 0.07 & 32 \\
& \verb|Karpathos|     & 0.18 & 0.12 & 0.15 & 24 \\
& \verb|Corfu|   & 0.08 & 0.04 & 0.05 & 27 \\
& \verb|Crete|        & 0.06 & 0.07 & 0.07 & 27 \\
& \verb|Cyprus|       & 0.04 & 0.06 & 0.05 & 18 \\
& \verb|Lesbos|       & 0.32 & 0.43 & 0.37 & 23 \\
& \verb|Laconia|      & 0.07 & 0.04 & 0.05 & 24 \\
& \verb|Macedonia|    & 0.00 & 0.00 & 0.00 & 20 \\
& \verb|Asia Minor|   & 0.04 & 0.05 & 0.04 & 22 \\
& \verb|Naxos|        & 0.00 & 0.00 & 0.00 & 19 \\
& \verb|Pontus|       & 0.24 & 0.30 & 0.26 & 30 \\
& \verb|Rhodes|        & 0.15 & 0.23 & 0.18 & 22 \\
& \verb|Skyros|       & 0.10 & 0.16 & 0.12 & 25 \\
\midrule
& \verb|accuracy|     & \multicolumn{1}{c}{} & \multicolumn{1}{c}{} & 0.13 & 575 \\
& \verb|macro avg|    & 0.12 & 0.13 & 0.12 & 575 \\
& \verb|weighted avg| & 0.13 & 0.13 & 0.12 & 575 \\
\bottomrule
\end{tabular}
\caption{Location classification with logistic regression using normalized data}
\end{table}

\begin{table}[h]
\centering
\setlength{\tabcolsep}{2pt}
\begin{tabular}{clcccc}
\toprule
& \textbf{Model} & \textbf{precision} & \textbf{recall} & \textbf{f1-score} & \textbf{support}\\
\midrule
& \verb|Epirus|       & 0.09 & 0.09 & 0.09 & 23 \\
& \verb|Aetolia|       & 0.42 & 0.46 & 0.44 & 24 \\
& \verb|Amorgos|       & 0.26 & 0.32 & 0.29 & 22 \\
& \verb|Eastern Thrace| & 0.19 & 0.25 & 0.22 & 24 \\
& \verb|Arcadia|       & 0.11 & 0.10 & 0.10 & 31 \\
& \verb|Achaea|         & 0.31 & 0.25 & 0.28 & 32 \\
& \verb|Ionian Islands|     & 0.47 & 0.70 & 0.56 & 23 \\
& \verb|Euboea|       & 0.06 & 0.05 & 0.05 & 20 \\
& \verb|Thesprotia|     & 0.11 & 0.09 & 0.10 & 22 \\
& \verb|Thrace|        & 0.26 & 0.20 & 0.23 & 25 \\
& \verb|Ioannina|     & 0.26 & 0.17 & 0.21 & 29 \\
& \verb|Karpathos|     & 0.42 & 0.39 & 0.41 & 28 \\
& \verb|Corfu|   & 0.25 & 0.22 & 0.24 & 27 \\
& \verb|Crete|        & 0.36 & 0.33 & 0.34 & 30 \\
& \verb|Cyprus|       & 0.70 & 0.96 & 0.81 & 24 \\
& \verb|Lesbos|       & 0.45 & 0.54 & 0.49 & 24 \\
& \verb|Laconia|      & 0.10 & 0.07 & 0.09 & 27 \\
& \verb|Macedonia|    & 0.35 & 0.30 & 0.32 & 27 \\
& \verb|Asia Minor|   & 0.20 & 0.11 & 0.14 & 18 \\
& \verb|Naxos|        & 0.44 & 0.58 & 0.50 & 24 \\
& \verb|Pontus|       & 0.73 & 0.84 & 0.78 & 19 \\
& \verb|Rhodes|        & 0.28 & 0.32 & 0.30 & 22 \\
& \verb|Skyros|       & 0.54 & 0.63 & 0.58 & 30 \\
\midrule
& \verb|accuracy|     & \multicolumn{1}{c}{} & \multicolumn{1}{c}{} & 0.34 & 575 \\
& \verb|macro avg|    & 0.32 & 0.35 & 0.33 & 575 \\
& \verb|weighted avg| & 0.32 & 0.34 & 0.33 & 575 \\
\bottomrule
\end{tabular}
\caption{Location classification with SVM using dialectal data}
\end{table}

\begin{table}[h]
\centering
\setlength{\tabcolsep}{2pt}
\begin{tabular}{clcccc}
\toprule
& \textbf{Model} & \textbf{precision} & \textbf{recall} & \textbf{f1-score} & \textbf{support}\\
\midrule
& \verb|Epirus|       & 0.05 & 0.04 & 0.05 & 23 \\
& \verb|Aetolia|      & 0.24 & 0.17 & 0.20 & 24 \\
& \verb|Amorgos|      & 0.11 & 0.07 & 0.08 & 29 \\
& \verb|Eastern Thrace| & 0.08 & 0.09 & 0.09 & 22 \\
& \verb|Arcadia|      & 0.09 & 0.07 & 0.08 & 28 \\
& \verb|Achaea|       & 0.26 & 0.19 & 0.22 & 27 \\
& \verb|Ionian Islands| & 0.24 & 0.20 & 0.22 & 30 \\
& \verb|Euboea|       & 0.15 & 0.17 & 0.16 & 24 \\
& \verb|Thesprotia|   & 0.16 & 0.25 & 0.19 & 24 \\
& \verb|Thrace|       & 0.05 & 0.03 & 0.04 & 31 \\
& \verb|Ioannina|     & 0.11 & 0.06 & 0.08 & 32 \\
& \verb|Karpathos|    & 0.19 & 0.17 & 0.18 & 24 \\
& \verb|Corfu|        & 0.16 & 0.11 & 0.13 & 27 \\
& \verb|Crete|        & 0.04 & 0.04 & 0.04 & 27 \\
& \verb|Cyprus|       & 0.06 & 0.06 & 0.06 & 18 \\
& \verb|Lesbos|       & 0.32 & 0.39 & 0.35 & 23 \\
& \verb|Laconia|      & 0.11 & 0.08 & 0.10 & 24 \\
& \verb|Macedonia|    & 0.00 & 0.00 & 0.00 & 20 \\
& \verb|Asia Minor|   & 0.00 & 0.00 & 0.00 & 22 \\
& \verb|Naxos|        & 0.06 & 0.11 & 0.07 & 19 \\
& \verb|Pontus|       & 0.28 & 0.33 & 0.30 & 30 \\
& \verb|Rhodes|       & 0.14 & 0.23 & 0.17 & 22 \\
& \verb|Skyros|       & 0.12 & 0.16 & 0.14 & 25 \\
\midrule
& \verb|accuracy|     & \multicolumn{1}{c}{} & \multicolumn{1}{c}{} & 0.13 & 575 \\
& \verb|macro avg|    & 0.13 & 0.13 & 0.13 & 575 \\
& \verb|weighted avg| & 0.13 & 0.13 & 0.13 & 575 \\
\bottomrule
\end{tabular}
\caption{Location classification with SVM using normalized data}
\end{table}

\begin{table}[h]
\centering
\setlength{\tabcolsep}{2pt}
\begin{tabular}{clcccc}
\toprule
& \textbf{Model} & \textbf{precision} & \textbf{recall} & \textbf{f1-score} & \textbf{support}\\
\midrule
& \verb|Epirus|       & 0.05 & 0.04 & 0.05 & 23 \\
& \verb|Aetolia|       & 0.26 & 0.29 & 0.27 & 24 \\
& \verb|Amorgos|       & 0.19 & 0.27 & 0.22 & 22 \\
& \verb|Eastern Thrace| & 0.14 & 0.21 & 0.17 & 24 \\
& \verb|Arcadia|       & 0.14 & 0.13 & 0.13 & 31 \\
& \verb|Achaea|         & 0.21 & 0.19 & 0.20 & 32 \\
& \verb|Ionian Islands|     & 0.28 & 0.57 & 0.38 & 23 \\
& \verb|Euboea|       & 0.06 & 0.05 & 0.05 & 20 \\
& \verb|Thesprotia|     & 0.06 & 0.05 & 0.05 & 22 \\
& \verb|Thrace|        & 0.27 & 0.16 & 0.20 & 25 \\
& \verb|Ioannina|     & 0.13 & 0.07 & 0.09 & 29 \\
& \verb|Karpathos|     & 0.38 & 0.21 & 0.27 & 28 \\
& \verb|Corfu|   & 0.18 & 0.19 & 0.18 & 27 \\
& \verb|Crete|        & 0.24 & 0.20 & 0.22 & 30 \\
& \verb|Cyprus|       & 0.53 & 0.71 & 0.61 & 24 \\
& \verb|Lesbos|       & 0.38 & 0.46 & 0.42 & 24 \\
& \verb|Laconia|      & 0.12 & 0.11 & 0.12 & 27 \\
& \verb|Macedonia|    & 0.24 & 0.15 & 0.18 & 27 \\
& \verb|Asia Minor|   & 0.00 & 0.00 & 0.00 & 18 \\
& \verb|Naxos|        & 0.25 & 0.29 & 0.27 & 24 \\
& \verb|Pontus|       & 0.57 & 0.68 & 0.62 & 19 \\
& \verb|Rhodes|        & 0.21 & 0.18 & 0.20 & 22 \\
& \verb|Skyros|       & 0.52 & 0.50 & 0.51 & 30 \\
\midrule
& \verb|accuracy|     & \multicolumn{1}{c}{} & \multicolumn{1}{c}{} & 0.25 & 575 \\
& \verb|macro avg|    & 0.23 & 0.25 & 0.23 & 575 \\
& \verb|weighted avg| & 0.24 & 0.25 & 0.23 & 575 \\
\bottomrule
\end{tabular}
\caption{Location classification with KNN using dialectal data}
\end{table}

\begin{table}[h]
\centering
\setlength{\tabcolsep}{2pt}
\begin{tabular}{clcccc}
\toprule
& \textbf{Model} & \textbf{precision} & \textbf{recall} & \textbf{f1-score} & \textbf{support}\\
\midrule
& \verb|Epirus|       & 0.06 & 0.09 & 0.07 & 23 \\
& \verb|Aetolia|      & 0.18 & 0.12 & 0.15 & 24 \\
& \verb|Amorgos|      & 0.08 & 0.07 & 0.08 & 29 \\
& \verb|Eastern Thrace| & 0.06 & 0.09 & 0.07 & 22 \\
& \verb|Arcadia|      & 0.15 & 0.11 & 0.12 & 28 \\
& \verb|Achaea|       & 0.12 & 0.07 & 0.09 & 27 \\
& \verb|Ionian Islands| & 0.36 & 0.13 & 0.20 & 30 \\
& \verb|Euboea|       & 0.12 & 0.17 & 0.14 & 24 \\
& \verb|Thesprotia|   & 0.23 & 0.29 & 0.25 & 24 \\
& \verb|Thrace|       & 0.04 & 0.03 & 0.03 & 31 \\
& \verb|Ioannina|     & 0.11 & 0.09 & 0.10 & 32 \\
& \verb|Karpathos|    & 0.18 & 0.17 & 0.17 & 24 \\
& \verb|Corfu|        & 0.04 & 0.04 & 0.04 & 27 \\
& \verb|Crete|        & 0.12 & 0.11 & 0.12 & 27 \\
& \verb|Cyprus|       & 0.06 & 0.06 & 0.06 & 18 \\
& \verb|Lesbos|       & 0.19 & 0.26 & 0.22 & 23 \\
& \verb|Laconia|      & 0.00 & 0.00 & 0.00 & 24 \\
& \verb|Macedonia|    & 0.06 & 0.05 & 0.05 & 20 \\
& \verb|Asia Minor|   & 0.00 & 0.00 & 0.00 & 22 \\
& \verb|Naxos|        & 0.06 & 0.11 & 0.07 & 19 \\
& \verb|Pontus|       & 0.25 & 0.20 & 0.22 & 30 \\
& \verb|Rhodes|       & 0.20 & 0.27 & 0.23 & 22 \\
& \verb|Skyros|       & 0.09 & 0.08 & 0.08 & 25 \\
\midrule
& \verb|accuracy|     & \multicolumn{1}{c}{} & \multicolumn{1}{c}{} & 0.11 & 575 \\
& \verb|macro avg|    & 0.12 & 0.11 & 0.11 & 575 \\
& \verb|weighted avg| & 0.12 & 0.11 & 0.11 & 575 \\
\bottomrule
\end{tabular}
\caption{Location classification with KNN using normalized data}
\end{table}

\begin{table}[h]
\centering
\setlength{\tabcolsep}{2pt}
\begin{tabular}{clcccc}
\toprule
& \textbf{Model} & \textbf{precision} & \textbf{recall} & \textbf{f1-score} & \textbf{support}\\
\midrule
& \verb|Epirus|       & 0.07 & 0.04 & 0.05 & 23 \\
& \verb|Aetolia|       & 0.33 & 0.71 & 0.45 & 24 \\
& \verb|Amorgos|       & 0.08 & 0.14 & 0.10 & 22 \\
& \verb|Eastern Thrace| & 0.15 & 0.21 & 0.17 & 24 \\
& \verb|Arcadia|       & 0.18 & 0.16 & 0.17 & 31 \\
& \verb|Achaea|         & 0.48 & 0.38 & 0.42 & 32 \\
& \verb|Ionian Islands|     & 0.24 & 0.22 & 0.23 & 23 \\
& \verb|Euboea|       & 0.00 & 0.00 & 0.00 & 20 \\
& \verb|Thesprotia|     & 0.13 & 0.14 & 0.13 & 22 \\
& \verb|Thrace|        & 0.43 & 0.24 & 0.31 & 25 \\
& \verb|Ioannina|     & 0.10 & 0.07 & 0.08 & 29 \\
& \verb|Karpathos|     & 0.58 & 0.25 & 0.35 & 28 \\
& \verb|Corfu|   & 0.21 & 0.22 & 0.21 & 27 \\
& \verb|Crete|        & 0.33 & 0.17 & 0.22 & 30 \\
& \verb|Cyprus|       & 0.55 & 0.88 & 0.68 & 24 \\
& \verb|Lesbos|       & 0.43 & 0.62 & 0.51 & 24 \\
& \verb|Laconia|      & 0.09 & 0.07 & 0.08 & 27 \\
& \verb|Macedonia|    & 0.11 & 0.04 & 0.06 & 27 \\
& \verb|Asia Minor|   & 0.00 & 0.00 & 0.00 & 18 \\
& \verb|Naxos|        & 0.35 & 0.38 & 0.36 & 24 \\
& \verb|Pontus|       & 0.40 & 0.74 & 0.52 & 19 \\
& \verb|Rhodes|        & 0.19 & 0.23 & 0.21 & 22 \\
& \verb|Skyros|       & 0.43 & 0.67 & 0.53 & 30 \\
\midrule
& \verb|accuracy|     & \multicolumn{1}{c}{} & \multicolumn{1}{c}{} & 0.29 & 575 \\
& \verb|macro avg|    & 0.25 & 0.28 & 0.25 & 575 \\
& \verb|weighted avg| & 0.26 & 0.29 & 0.26 & 575 \\
\bottomrule
\end{tabular}
\caption{Location classification with Random Forest using dialectal data}
\end{table}

\begin{table}[h]
\centering
\setlength{\tabcolsep}{2pt}
\begin{tabular}{clcccc}
\toprule
& \textbf{Model} & \textbf{precision} & \textbf{recall} & \textbf{f1-score} & \textbf{support}\\
\midrule
& \verb|Epirus|       & 0.06 & 0.09 & 0.07 & 23 \\
& \verb|Aetolia|      & 0.07 & 0.04 & 0.05 & 24 \\
& \verb|Amorgos|      & 0.31 & 0.14 & 0.19 & 29 \\
& \verb|Eastern Thrace| & 0.15 & 0.14 & 0.14 & 22 \\
& \verb|Arcadia|      & 0.04 & 0.04 & 0.04 & 28 \\
& \verb|Achaea|       & 0.30 & 0.22 & 0.26 & 27 \\
& \verb|Ionian Islands| & 0.24 & 0.30 & 0.27 & 30 \\
& \verb|Euboea|       & 0.11 & 0.12 & 0.12 & 24 \\
& \verb|Thesprotia|   & 0.18 & 0.17 & 0.17 & 24 \\
& \verb|Thrace|       & 0.10 & 0.06 & 0.08 & 31 \\
& \verb|Ioannina|     & 0.12 & 0.06 & 0.08 & 32 \\
& \verb|Karpathos|    & 0.15 & 0.12 & 0.14 & 24 \\
& \verb|Corfu|        & 0.07 & 0.04 & 0.05 & 27 \\
& \verb|Crete|        & 0.00 & 0.00 & 0.00 & 27 \\
& \verb|Cyprus|       & 0.03 & 0.06 & 0.04 & 18 \\
& \verb|Lesbos|       & 0.35 & 0.35 & 0.35 & 23 \\
& \verb|Laconia|      & 0.00 & 0.00 & 0.00 & 24 \\
& \verb|Macedonia|    & 0.06 & 0.10 & 0.07 & 20 \\
& \verb|Asia Minor|   & 0.03 & 0.05 & 0.04 & 22 \\
& \verb|Naxos|        & 0.05 & 0.05 & 0.05 & 19 \\
& \verb|Pontus|       & 0.27 & 0.40 & 0.32 & 30 \\
& \verb|Rhodes|       & 0.17 & 0.32 & 0.22 & 22 \\
& \verb|Skyros|       & 0.12 & 0.16 & 0.14 & 25 \\
\midrule
& \verb|accuracy|     & \multicolumn{1}{c}{} & \multicolumn{1}{c}{} & 0.13 & 575 \\
& \verb|macro avg|    & 0.13 & 0.13 & 0.13 & 575 \\
& \verb|weighted avg| & 0.13 & 0.13 & 0.13 & 575 \\
\bottomrule
\end{tabular}
\caption{Location classification with Random Forest using normalized data}
\end{table}

\clearpage

\begin{table}[h]
\centering
\setlength{\tabcolsep}{2pt}
\begin{tabular}{clcccc}
\toprule
& \textbf{Model} & \textbf{lat MAE} & \textbf{lon MAE} & \textbf{lat MSE} & \textbf{lon MSE}\\
\midrule
&\verb|ElasticNet| & {\textbf{1.37}} & {\textbf{2.77}} & {\textbf{2.94}} & {\textbf{14.31}} \\
&\verb|K Nearest Neighbors| & {1.47} & {3.13} & {3.34} & {16.65} \\
&\verb|Linear Regression| & {1.38} & {2.80} & {3.00} & {14.70} \\
&\verb|Random Forest| & {1.43} & {2.82} & {3.16} & {14.63} \\
&\verb|Extremely Randomized Trees| & {1.43} & {2.84} & {3.15} & {14.68} \\
\bottomrule
\end{tabular}
\caption{Geolocation regression using dialectal data}
\end{table}

\begin{table}[h]
\centering
\setlength{\tabcolsep}{2pt}
\begin{tabular}{clcccc}
\toprule
& \textbf{Model} & \textbf{lat MAE} & \textbf{lon MAE} & \textbf{lat MSE} & \textbf{lon MSE}\\
\midrule
&\verb|ElasticNet| & {1.51} & {2.98} & {3.40} & {17.68} \\
&\verb|K Nearest Neighbors| & {1.55} & {2.96} & {3.57} & {17.47} \\
&\verb|Linear Regression| & {1.54} & {3.08} & {3.57} & {18.44} \\
&\verb|Random Forest| & {1.51} & {2.90} & {3.42} & {17.18} \\
&\verb|Extremely Randomized Trees| & {1.52} & {2.92} & {3.47} & {17.40} \\
&\verb|GreekBERT| & {\textbf{1.35}} & {\textbf{1.83}} & {\textbf{2.76}} & {\textbf{5.57}} \\
\bottomrule
\end{tabular}
\caption{Geolocation regression using normalized data}
\end{table}

\FloatBarrier
\section{GreekBERT Fine-Tuning Hyperparameters}
\label{sec:hyper}
We add a 30\% dropout and a single linear layer as a regressor on top of the Greek BERT model and train it on 80\% of out data, keeping 10\% as a validation set for early stopping after 2 epochs of non-improvement, for a maximum of 15 epochs. We then test it on the remaining 10\% of our data. We use mean squared error as the loss function, AdamW as the optimizer, \(2\times10^{-5}\) as the learning rate and a batch size of 32.

\end{document}